\documentclass[11pt,a4paper]{article}
\usepackage[hyperref]{acl2019}
\usepackage{times}
\usepackage{latexsym}

\usepackage{url}
\usepackage{amsmath}
\usepackage{graphicx}
\usepackage{xcolor}
\usepackage{subcaption}

\usepackage{amsmath,amsfonts,bm}

\def\eqref#1{equation~\ref{#1}}

\def\1{\bm{1}}

\def\va{{\bm{a}}}
\def\vb{{\bm{b}}}

\def\ve{{\bm{e}}}

\def\vg{{\bm{g}}}
\def\vh{{\bm{h}}}

\def\vm{{\bm{m}}}

\def\vo{{\bm{o}}}

\def\vv{{\bm{v}}}

\def\vx{{\bm{x}}}

\def\mA{{\bm{A}}}

\def\mK{{\bm{K}}}

\def\mM{{\bm{M}}}

\def\mQ{{\bm{Q}}}

\def\mS{{\bm{S}}}

\def\mV{{\bm{V}}}
\def\mW{{\bm{W}}}
\def\mX{{\bm{X}}}

\DeclareMathAlphabet{\mathsfit}{\encodingdefault}{\sfdefault}{m}{sl}
\SetMathAlphabet{\mathsfit}{bold}{\encodingdefault}{\sfdefault}{bx}{n}

\DeclareMathOperator*{\argmax}{arg\,max}

\aclfinalcopy 

\title{Episodic Memory Reader: Learning What to Remember\\ for Question Answering from Streaming Data}

\author{Moonsu Han$^1$$^*$\thanks{* Equal contribution} \quad\quad Minki Kang$^1$$^*$ \quad\quad Hyunwoo Jung$^1$ \quad\quad Sung Ju Hwang$^1$$^,$$^2$ \\  \\
	KAIST$^1$, Daejeon, South Korea \\ AITRICS$^2$, Seoul, South Korea \\
	\texttt{\{mshan92, zzxc1133, hyunwooj, sjhwang82\}@kaist.ac.kr}}
\renewcommand\footnotemark{}

\date{}

\begin{document}
\maketitle
\begin{abstract}

We consider a novel question answering (QA) task where the machine needs to read from large streaming data (long documents or videos) without knowing when the questions will be given, which is difficult to solve with existing QA methods due to their lack of scalability. To tackle this problem, we propose a novel end-to-end deep network model for reading comprehension, which we refer to as \textbf{E}pisodic \textbf{M}emory \textbf{R}eader (EMR) that sequentially reads the input contexts into an external memory, while replacing memories that are less important for answering \emph{unseen} questions. Specifically, we train an RL agent to replace a memory entry when the memory is full, in order to maximize its QA accuracy at a future timepoint, while encoding the external memory using either the GRU or the Transformer architecture to learn representations that considers relative importance between the memory entries. We validate our model on a synthetic dataset (bAbI) as well as real-world large-scale textual QA (TriviaQA) and video QA (TVQA) datasets, on which it achieves significant improvements over rule-based memory scheduling policies or an RL-based baseline that independently learns the query-specific importance of each memory.
\end{abstract}

\begin{figure*}[ht]
	\centering
	{\includegraphics[width=0.9\linewidth]{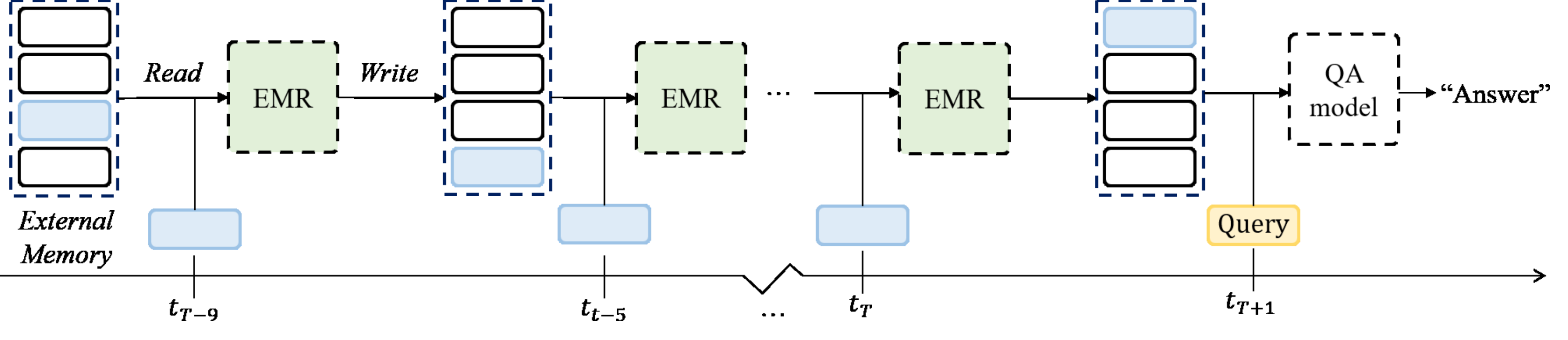}}
	\vspace{-0.1in}
	\caption{\small \textbf{Concept:} We consider a novel problem of learning from streaming data, where the QA model may need to answer a question that is given after reading in unlimited amount of context. To solve this problem, our Episodic Memory Reader (EMR) learns to retain the most important context vectors in an external memory, while replacing the memory entries in order to maximize its accuracy on an unseen question given at a future timestep.}
	\label{fig:conceptfigure}
	\centering
	\vspace{-0.40cm}
\end{figure*}

\section{Introduction}
Question answering (QA) problem is one of the most important challenges in Natural Language Understanding (NLU). In recent years, there has been drastic progress on the topic, owing to the success of deep learning based QA models~\citep{MemN2N, BiDAF,  DMNP, MnemonicReader, MemoReader, BERT}. On certain tasks such as machine reading comprehension (MRC), where the problem is to find the span of the answer within a given paragraph~\cite{SQuAD}, the deep-learning based QA models have even surpassed human-level performances. 

Despite such impressive achievements, it is still challenging to model question answering with document-level context~\cite{TriviaQA}, where the context may include a long document with a large number of paragraphs, due to problems such as difficulty in modeling long-term dependency and computational cost. To overcome such scalability problems, researchers have proposed pipelining or confidence based selection methods that combine paragraph-level models to obtain a document-level model~\citep{TriviaQA, ReadWiki, DocumentQA, R3}. Yet, such models are applicable only when questions are given beforehand and all sentences in the document can be stored in memory.

However, in realistic settings, the amount of context may be too large to fit into the system memory. We may consider query-based context selection methods such as ones proposed in~\citet{CuttoChase} and~\citet{MinimalContext}, but in many cases, the question may not be given when reading in the context, and thus it would be difficult to select out the context based on the question. For example, a conversation agent may need to answer a question after numerous conversations in a long-term time period, and a video QA model may need to watch an entire movie, or a sports game, or days of streaming videos from security cameras before answering a question. In such cases, existing QA models will fail to solve the problem due to memory limitation. 

In this paper, we target a novel problem of solving question answering problem with streaming data as context, where the size of the context could be significantly larger than what the memory can accommodate (See Figure~\ref{fig:conceptfigure}). In such a case, the model needs to carefully manage what to remember from this streaming data such that the memory contains the most informative context instances in order to answer an unseen question in the future. We pose this memory management problem as a learning problem and train both the memory representation and the scheduling agent using reinforcement learning. 

Specifically, we propose to train the memory module itself using reinforcement learning to replace the most uninformative memory entry in order to maximize its reward on a given task. However, this is a seemingly ill-posed problem since for most of the time, the scheduling should be performed without knowing which question will arrive next. To tackle this challenge, we implement the policy network and the value network that learn not only relation between sentences and query but also relative importance among the sentences in order to maximize its question answering accuracy at a future timepoint. We refer to this network as Episodic Memory Reader (EMR). EMR can perform selective memorization to keep a compact set of important context that will be useful for future tasks in lifelong learning scenarios. 

We validate our proposed memory network on a large-scale QA task (TriviaQA) and video question answering task (TVQA) where the context is too large to fit into the external memory, against rule-based and an RL-based scheduling method without consideration of relative importance between memories. The results show that our model significantly outperforms the baselines, due to its ability to preserve the most important pieces of information from the streaming data.

Our contribution is threefold:
\begin{itemize}
	\item We consider a novel task of learning to remember important instances from streaming data for question answering task, where the size of the memory is significantly smaller than the length of the data stream. 
	\item We propose a novel end-to-end memory-augmented neural architecture for solving QA from streaming data, where we train a scheduling agent via reinforcement learning to store the most important memory entries for solving future QA tasks.
	\item We validate the efficacy of our model on real-world large-scale text and video QA datasets, on which it obtains significantly improved performances over baseline methods. 
\end{itemize}
\section{Related Work}

\paragraph{Question-answering} There has been a rapid progress in question answering (QA) in recent years, thanks to the advancement in deep learning as well as the availability of large-scale datasets. One of the most popular large-scale QA dataset is Stanford Question Answering Dataset (SQuAD)  \cite{SQuAD} that contains 100K question-answering pairs. Unlike \citet{MCTest} and \citet{CNN/Daily} that provide multiple-choice QA pairs, SQuAD provides and requires to predict exact locations of the answers. On this span prediction task, attentional models~\citep{MEMEN, AoAReader, MnemonicReader} have achieved impressive performances, with Bi-Directional Attention Flow (BiDAF) \cite{BiDAF} that uses bi-directional attention mechanism for the context and query being one of the best performing models. TriviaQA \cite{TriviaQA} is another large-scale QA dataset that includes $950K$ QA pairs. Since the length of each document in Trivia is much longer than SQuAD, with average of $3K$ sentences per document, existing span prediction models~\citep{TriviaQA, MemoReader, QANet} fail to work due to memory limitation, and simply resort to document truncation. Video question answering~\citep{MovieQA, TVQA}, where video frames are given as context for QA, is another important topic where scalability is an issue. Several models~\citep{DeepStroy,MDAM, RWMN, LMN} propose to solve video QA using attentions and memory augmented networks, to perform composite reasoning over both videos and texts; however, they only focus on short-length videos. Most existing work on QA focus on small-size problems due to memory limitation. Our work, on the other hand, considers a challenging scenario where the context is order of magnitude larger than the memory.

\paragraph{Context selection} A few recent models propose to select minimal context from the given document when answering questions for scalability, rather than using the full context. \citet{MinimalContext} proposed a context selector that generates attentions on the context vectors, in order to achieve scability and robustness against adversarial inputs. \citet{CoarsetoFine} and \citet{CuttoChase} propose a similar method, but they use REINFORCE~\cite{REINFORCE} instead of linear classifiers. \citet{ReadWiki} selects the most relevant documents out of the Wikipedia database with respect to the query using TF-IDF matching, and \citet{R3} propose to tackle the document ranking problem with RL agents. While these context/document selection methods share our motivation of achieving scability and selecting out the most informative pieces of information to solve the QA task, our problem setting is completely different from theirs since we consider a challenging problem of learning from the streaming data \emph{without knowing} when the question will be given, where the size of the context is much larger than the memory and the question is unseen when training the selection module. 

\paragraph{Memory-augmented neural networks} Our episodic memory reader is essentially a memory-augmented network (MANN)~\citep{NTM, MemN2N, DMNP, DMN} with a RL-based scheduler. While most existing work on MANN assume that the memory is sufficiently large to hold all the data instances, a few tried to consider memory-scheduling for better scalability. \citet{DNTM} propose to train an addressing agent using reinforcement learning in order to dynamically decide which memory to overwrite based on the query. This query-specific importance is similar to our motivation, but in our case the query is given after reading in all the context and thus unusable for scheduling, and we perform hard replacement instead of overwriting. Differentiable Neural Computer (DNC)~\cite{DNC} extends the NTM to address the issue by introducing a temporal link matrix, replacing the least used memory when the memory is full. However, this method is a rule-based one that cannot maximize the performance on a given task.

\begin{figure*}[ht]
	\centering
	{\includegraphics[width=0.9\linewidth]{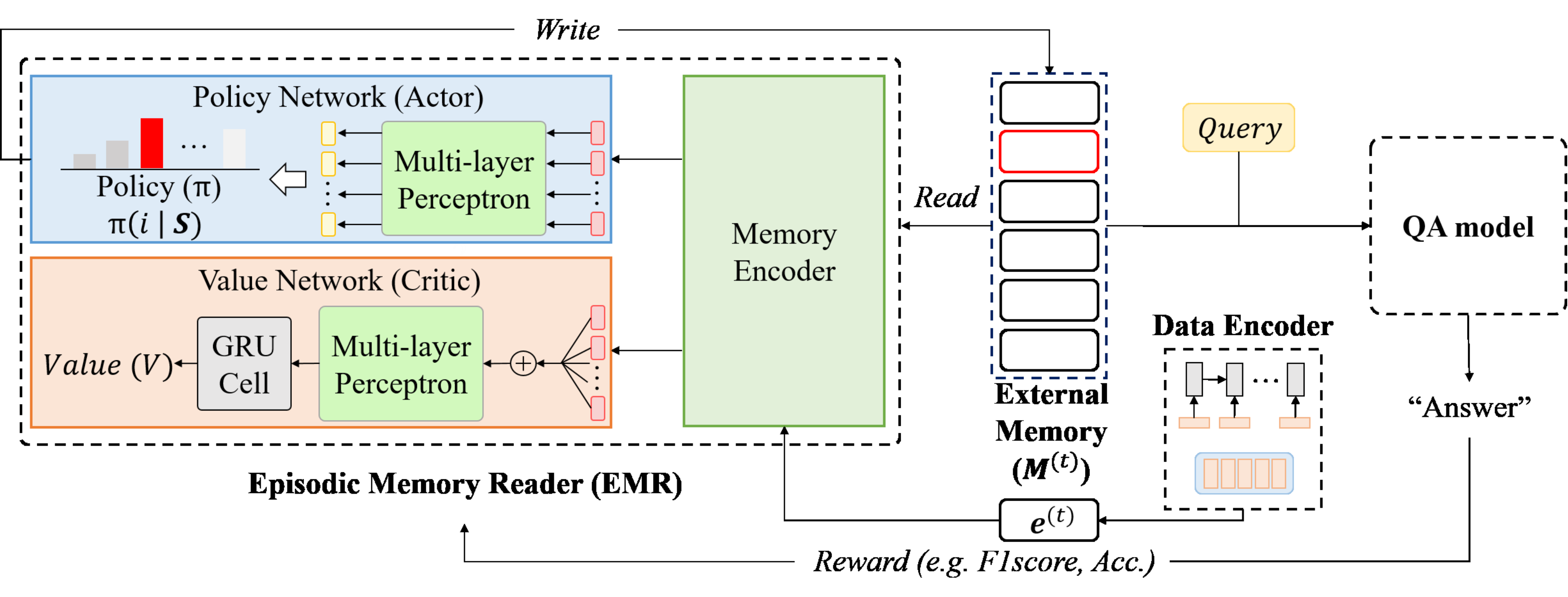}}
	\caption{\small The overview of our Episodic Memory Reader (EMR). EMR learns the policy and the value network to select a memory entry to replace, in order to maximize the reward, defined as the performance on future QA tasks (F1-score, accuracy).}
	\label{fig:fullarchitecture}
	\centering
	\vspace{-0.5cm}
\end{figure*}

\section{Learning What to Remember from Streaming Data}
We now describe how to solve question answering tasks with streaming data as context. In a more general sense, this is a problem of learning from a long data stream that contains a large portion of unimportant, noisy data (e.g. routine greetings in dialogs, uninformative video frames) with limited memory. The data stream is episodic, where an unlimited amount of data instances may arrive at one time interval and becomes inaccessible afterward. Additionally, we consider that it is not possible for the model to know in advance what tasks (a question in the case of QA problem) will be given at which timestep in the future (See Figure~\ref{fig:conceptfigure} for more details). To solve this problem, the model needs to identify important data instances from the data stream and store them into external memory. Formally, given a data stream (e.g. sentences or images) $\mX = \{ \vx^{(1)}, \cdots, \vx^{(T)} \}$ where $\vx^{(t)}\in\mathbb{R}^d$ as input, the model should learn a function $\mathcal{F} : \mX\mapsto{\mM}$ that maps it to the set of memory entries $\mM = \{\vm_1, \cdots, \vm_N\}$ where $\vm_i\in\mathbb{R}^k$ and $T \gg N$. How can we then learn such a function that maximizes the performance on unseen future tasks without knowing what problems will be given at what time? We formulate this problem as a reinforcement learning problem to train a memory scheduling agent.

\subsection{Model Overview}
We now describe our model, \emph{Episodic Memory Reader (EMR)} to solve the previously described problem. Our model has three components: (1) an agent $\mathcal{A}$ based on EMR, (2) an external memory $\mM = [\vm_1, \cdots, \vm_N]$, and (3) a solver which solves the given task (e.g. QA) with the external memory. Figure \ref{fig:fullarchitecture} shows the overview of our model. Basically, given a sequence of data instances $\mX = \{ \vx^{(1)}, \cdots, \vx^{(T)} \}$ that streams through the system, the agent learns to retain the most useful subset in the memory, by interacting with the external memory that encodes the relative importance of each memory entry. When $t \leq N$, the agent simply maps $\vx^{(t)}$ to $\vm^{(t)}$. However, when $t > N$, when the memory becomes full, it selects an existing memory entry to delete. Specifically, it outputs an action based on $\pi (i | \mS^{(t)})$, which denotes the selection of $i_{th}$ memory entry to delete. Here, the state is the concatenation of the memory and the data instance: $\mS^{(t)}=[\mM^{(t)}, \ve^{(t)}]$, where $\ve^{(t)}$ is the encoded input at timestep $t$. To maximize the performance on the future QA task, the agent should replace the least important memory entry. When the agent encounters the task $\mathcal{T}$ (QA problem) at timestep $T+1$, it leverages both the memory at timestep $T$, $\mM^{(T)}$ and the task information (e.g. question), to solve the task. For each action, the environment (QA module) provides the reward $\mathcal{R}^{(t)}$, that is given either as the F1-score or the accuracy.

\subsection{Episodic Memory Reader}
Episodic Memory Reader (EMR) is composed of three components: \textbf{(1) Data Encoder} that encodes each data instance into memory vector representation, \textbf{(2) Memory Encoder} that generates replacement probability for each memory entry, and the \textbf{(3) Value Network} that estimates the value of memory as a whole. In some cases, we may use policy gradient methods, in which case the value network becomes unnecessary.

\subsubsection{Data Encoder}
The data instance $\vx^{(t)}$ which arrives at time $t$ can be in any data format, and thus we transform it into a $k$-dimensional memory vector representation $\ve^{(t)} \in \mathbb{R}^k$ to  using an encoder:

\begin{equation*} \label{eq:mem_repr}
\ve^{(t)} = \psi (\vx^{(t)})
\end{equation*}

\noindent where $\psi (\cdot)$ is the data encoder, which could be any neural architecture based on the type of the input data. For example, we could use a RNN if $\vx^{(t)}$ is composed of sequential data (e.g. a sentence composed of words $\vx^{(t)} = \{w_1, w_2, w_3, \cdots, w_s\}$) or a CNN if $\vx^{(t)}$ is an image. After deleting a memory entry $\vm_i^{(t)}$, we append $\ve^{(t)}$ at the end of the memory, which then becomes $\vm^{(t+1)}_N$.


\subsubsection{Memory Encoder}
Using the memory vector representations $\mM^{(t)} = [\vm_{1}^{(t)}, \cdots, \vm_{N}^{(t)}]$ and $\ve^{(t)}$ generated from the data encoder, the memory encoder outputs a probability for each memory entry by considering its relative importance, and then replaces the most unimportant entry. This component corresponds to the policy network of the actor-critic method. Now we describe our EMR models.

\paragraph{EMR-Independent}
Since we do not have existing work for our novel problem setting, as a baseline, we first consider a memory encoder that only captures the relative importance of each memory entry independently to the new data instance, which we refer to as EMR-Independent. This scheduling mechanism is adopted from Dynamic Least Recently Use (LRU) addressing introduced in~\citet{DNTM}, but different from LRU in that it replaces the memory entry rather than overwriting it, and is trained without query to maximize the performance for unseen future queries. EMR-Independent outputs the importance for each memory entry by comparing them with an embedding of the new data instance $\vx^{(t)}$ as $\va^{(t)}_{i} = \text{softmax}(\vm^{(t)}_{i} \psi(\vx^{(t)})^T)$. To compute the overall importance of each memory entry, as done in \citet{DNTM}, we compute the exponential moving average as $\vv^{(t)}_{i} = 0.1 \vv^{(t-1)}_{i} + 0.9 \va^{(t)}_{i}$. Then, we compute the replacement probabilty of each memory entry with the LRU factor $\gamma^{(t)}$ as follows:

\begin{gather*}
\gamma^{(t)}_{i} = \sigma(\mW_{\gamma}^T \vm^{(t)}_{i} + b_{\gamma})\\
\vg^{(t)}_{i} = \va^{(t)}_{i} - \gamma^{(t)}_{i} \vv^{(t-1)}_{i}\\
\pi (i | [\mM^{(t)}, e^{(t)}];\theta) = \text{softmax}(\vg^{(t)}_{i})
\end{gather*}

\noindent where $i \in [1, N]$ is the memory index, $\mW_{\gamma} \in \mathbb{R}^{1 \times d}$ and $\vb_{\gamma} \in \mathbb{R}$ are the weight matrix and the bias term,  $\sigma(\cdot)$ and $\text{softmax}(\cdot)$ are sigmoid and softmax functions respectively, and $\pi$ is the policy of the memory scheduling agent.

\paragraph{EMR-biGRU}
A major drawback of EMR-Independent is that the evaluation of each memory depends only on the input $\vx^{(t)}$. In other words, the importance is computed between each memory entry and the new data instance regardless of other entries in the memory. However, this scheme cannot model the relative importance of each memory entry to other memory entries, which is more important in deciding on the least important memory. One way to consider relative relationships between memory entries is to encode them using a bidirectional GRU (biGRU) as follows:

\begin{gather*}
\overrightarrow{\vh}^{(t)}_{i} = GRU_{\theta_{fw}}(\vm^{(t)}_{i}, \overrightarrow{\vh}^{(t)}_{i-1})\\ 
\overleftarrow{\vh}^{(t)}_{i} = GRU_{\theta_{bw}}(\vm^{(t)}_{i}, \overleftarrow{\vh}^{(t)}_{i+1})\\
\vh^{(t)}_{i} = [\overrightarrow{\vh}^{(t)}_{i}, \overleftarrow{\vh}^{(t)}_{i}]\\
\pi (i | [\mM^{(t)}, \ve^{(t)}];\theta) = \text{softmax}(MLP(\vh_i^{(t)}))
\end{gather*}

\noindent where $i \in [1, N+1]$ is the memory index, including the index of the encoded input $\vm^{(t)}_{N+1} = \ve^{(t)}$, $GRU_\theta$ is a Gated Recurrent Unit parameterized by $\theta$, $[\overrightarrow{\vh}^{(t)}_{i}, \overleftarrow{\vh}^{(t)}_{i}]$ is a concatenation of features. $\pi$ is the policy of the agent, and MLP is a multi-layer perceptron with three layers with ReLU activation functions.
Thus, EMR-biGRU learns the general importance of each memory entry in relation to its neighbors rather than independently computing the importance of each entry with respect to the query, which is useful when selecting out the most important entries among highly similar data instances (e.g. video frames). However, the model may not effectively model long-range relationships between memory entries in far-away slots due to the inherent limitation with RNNs.

\begin{figure}[t]
	\centering
	{\includegraphics[width=\linewidth]{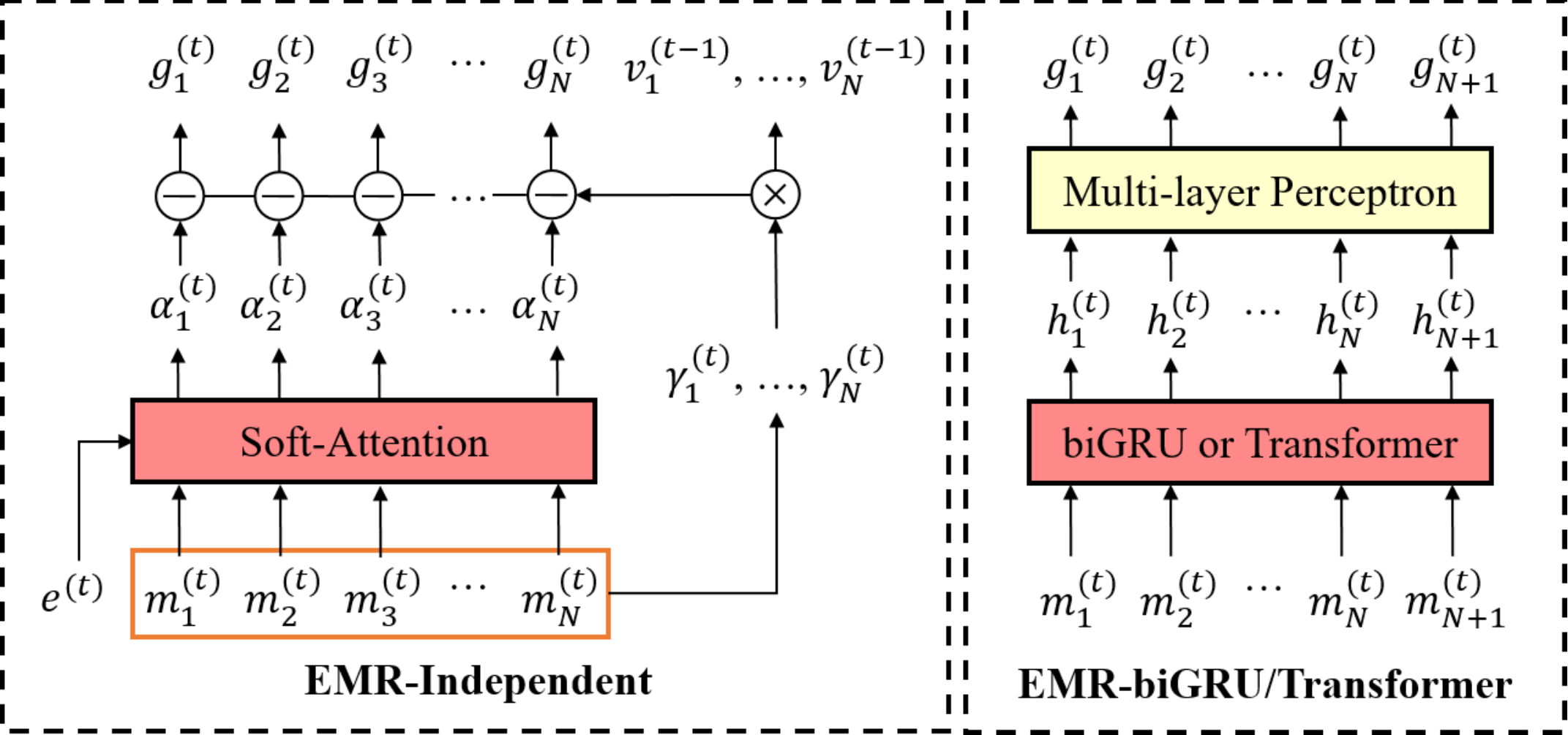}}
	\vspace{-0.50cm}
	\caption{\small Detailed architecture of memory encoder in EMR-Independent and EMR-biGRU/Transformer.}
	\label{fig:architecture}
	\centering
	\vspace{-0.50cm}
\end{figure}

\paragraph{EMR-Transformer} To overcome such suboptimality of RNN-based modeling, we further adopt the self-attention mechanism from~\citet{Self-Attention}. With query $\mQ^{(t)}$, key $\mK^{(t)}$, and the value $\mV^{(t)}$ we generate the relative importance of the entries with a linear layer that takes $\vm^{(t)}$ with the position encoding proposed in~\citet{Self-Attention} as input. With multi-headed attention, each component is projected to a multi-dimensional space; the dimensions for each componenets are $\mQ^{(t)} \in \mathbb{R}^{H \times N \times \frac{k}{H}}$, $\mK^{(t)} \in \mathbb{R}^{H \times N \times \frac{k}{H}}$, and $\mV^{(t)} \in \mathbb{R}^{H \times N \times \frac{k}{H}}$, where $N$ is the size of memory and $H$ is the number of attention heads. Using these, we can formulate the retrieved output using self-attention and memory encoding as follows:

\begin{gather*}
\mA^{(t)} = \text{softmax}\left( \frac{\mQ^{(t)}\mK^{(t)^T}}{\sqrt{k/H}}\right)\\
\vo^{(t)} = \mA^{(t)} \mV^{(t)}\\
\vh^{(t)} = \mW_o^T [\vo_1^{(t)}, \vo_2^{(t)}, \cdots, \vo_h^{(t)}]\\
\pi (i | [\mM^{(t)}, \ve^{(t)}];\theta) = \text{softmax}(MLP(\vh_i^{(t)}))
\end{gather*}

\noindent where $i$ is the memory index, $\vo_i^{(t)} \in \mathbb{R}^{N \times \frac{d}{h}}$, $[\vo_1^{(t)}, \vo_2^{(t)}, \cdots, \vo_h^{(t)}] \in \mathbb{R}^{N \times k}$ is a concatentation of $\vo_i^{(t)}$, $\pi$ is the policy of the agent, and $MLP$ is the same 3-layer multi-layer perceptron used in EMR-biGRU. Memory encoding $\vh^{(t)}$ is then computed using linear function $\mW_o \in \mathbb{R}^{d \times d}$ with $\vh^{(t)}$ as input. Figure~\ref{fig:architecture} illustrates the architecture of the memory encoder for EMR-Independent and EMR-biGRU/Transformer.

\subsubsection{Value Network}
For solving certain QA problems, we need to consider the future importance of each memory entry. Especially in textual QA tasks (e.g. TriviaQA), storing the evidence sentences that precede span words may be useful as they may provide useful context. However, using only discrete policy gradient method, we cannot preserve such context instances. To overcome this issue, we use an actor-critic RL method (A3C) \cite{A3C} to estimate the sum of future rewards at each state using the value network. The difference between the policy and the value is that the value can be estimated differently at each time step and the needs to consider the memory as a whole. To obtain a holistic representation of our memory, we use Deep Sets \cite{DeepSets}. Following \citet{DeepSets} we sum up all $\vh^{(t)}_{i}$ and input them into an MLP ($\rho$), that consists of two linear layers and a ReLU activation function, to obtain a set representation. Then, we further process the set representation $\rho(\sum_{i=1}^N \vh^{(t)}_i)$ by a GRU with the hidden state from the previous time step. Finally, we feed the output of the GRU to a multi-layer perceptron to estimate the value $V^{(t)}$ for the current timestep. 

\begin{figure}[t]
\includegraphics[width=\linewidth]{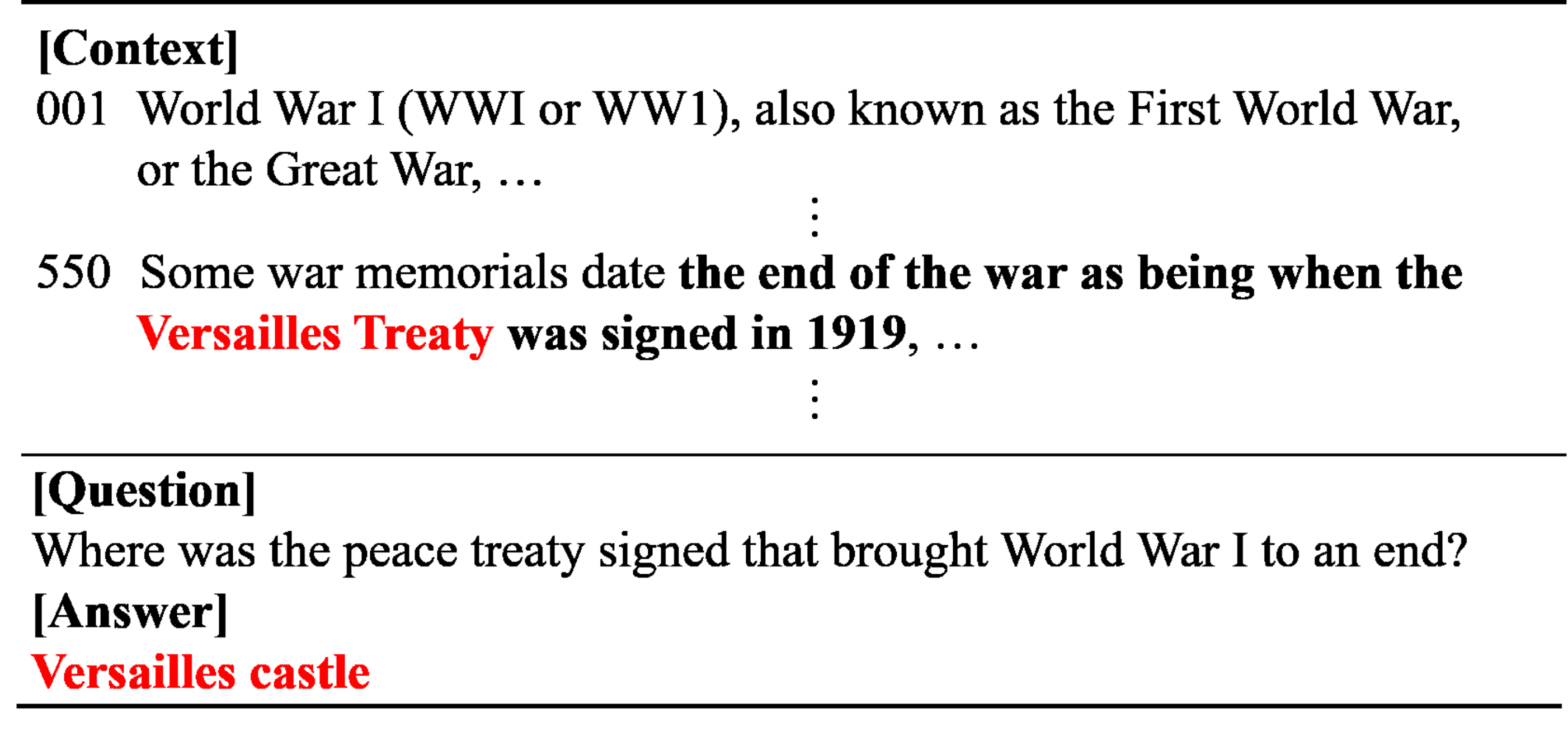} 
\vspace{-0.3in}
\caption{\small Example context and QA pair from TriviaQA.}
\label{fig:trivia/example}
\end{figure}

\subsection{Training and test}

\paragraph{Training}
Our model learns the memory scheduling policy jointly with the model to solve the task. For training EMR, we choose A3C \cite{A3C} or REINFORCE \cite{REINFORCE}. At training time, since the tasks are given, we provide the question to the agent at every timestep. At each step, the agent selects the action stochastically from multinomial distribution based on $\pi (i | [\mM^{(t)}, \ve^{(t)}] ; \theta)$ to explore various states, and make an action. Then, the QA model provides the agent the reward $\mathcal{R}_t$. We use asynchronous multiprocessing method illustrated in \citet{A3C} to train several models at once. 



\paragraph{Test}
At test time, the agent deletes the memory index by following the learned policy $\pi$: $\argmax_i\pi (i| [\mM^{(t)}, \ve^{(t)}] ; \theta)$. Contrarily from the training step, the model observes the question only at the end of the data stream. When encountering the question, the model solves the task using the data instances kept in the external memory. 

\section{Experiment}

We experiment our EMR-biGRU and EMR-Transformer against several baselines:

\textbf{1) FIFO (First-In First-Out).}
A rule-based memory scheduling policy that replaces the oldest memory entry. 

\textbf{2) Uniform.}
A policy that replaces all memory entries with equal probability at each time. 

\textbf{3) LIFO (Last-In First-Out).}
A policy that replaces the newest data instance. That is, it first fills in the memory and then discards all following data instances.

\textbf{4) EMR-Independent.}
A baseline EMR which learns the importance of each memory entry only relative to the new data instance.

\textbf{5) EMR-biGRU.}
An EMR implemented using a biGRU, that considers relative importance of each memory entry to its neighbors when learning the memory replacement policy.

\textbf{6) EMR-Transformer}.
An EMR that utilizes Transformer to model the global relative importance between memory entries.

The codes for the baseline models and our models are available at \url{https://github.com/h19920918/emr}. In the next subsections, we present experimental results on bAbI, TriviaQA, and TVQA datasets. For more experimental results, please see \textbf{supplementary file}.


\subsection{bAbI \label{exp_babi}}

\paragraph{Dataset}
bAbI \cite{bAbI} dataset, which is a synthetic dataset for episodic question answering, consists of 20 tasks with small amount of vocabulary, that can be solved by remembering a person or an object. Among the $20$ tasks, we select Task 2, which requires to remember two supporting facts, to evaluate our model. Additionally, we generate noisy version of this task using the open-source template provied by \citet{bAbI}. Each episode of both tasks contains five questions, where all questions share the same context sentences. For Noisy task, we inject noise sentences that has nothing to do with the given task, to validate the effitiveness of our model. In this dataset, 60\% of the episodes have no noise sentence, 10\% have approximately 30\% noise sentences, 10\% have approximately 45\% noise sentences and 10\% of dataset have approximately 60\% noise sentences. The length of each episode is fixed to 45 (5 questions + 40 facts), and a question appears after the arrival of 8 factual sentences. 

\paragraph{Experiment Details}
We adopted MemN2N \cite{MemN2N} for this experiment, which consists of an embedded layer and a multi-hop mechanism that extracts high-level inference. We use MemN2N with position encoding representation, 3 hops and adjacent weight tying. We set the dimension of memory representations to $k=20$ and compare our model and the baselines on the Original and Noisy tasks. To generate the memory representation $\vm_{i}$, we use the sum of the three hop value memories from the base MemN2N. We experiment with varying memory size: $5, 10$ and $15$. We train our model and the baselines using ADAM optimizer \cite{Adam} with the learning rate 0.0005 for 400K steps on both tasks.

\begin{figure}[t]
	\centering
	\begin{subfigure}[b]{0.23\textwidth}
		\centering
		\includegraphics[width=4.0cm]{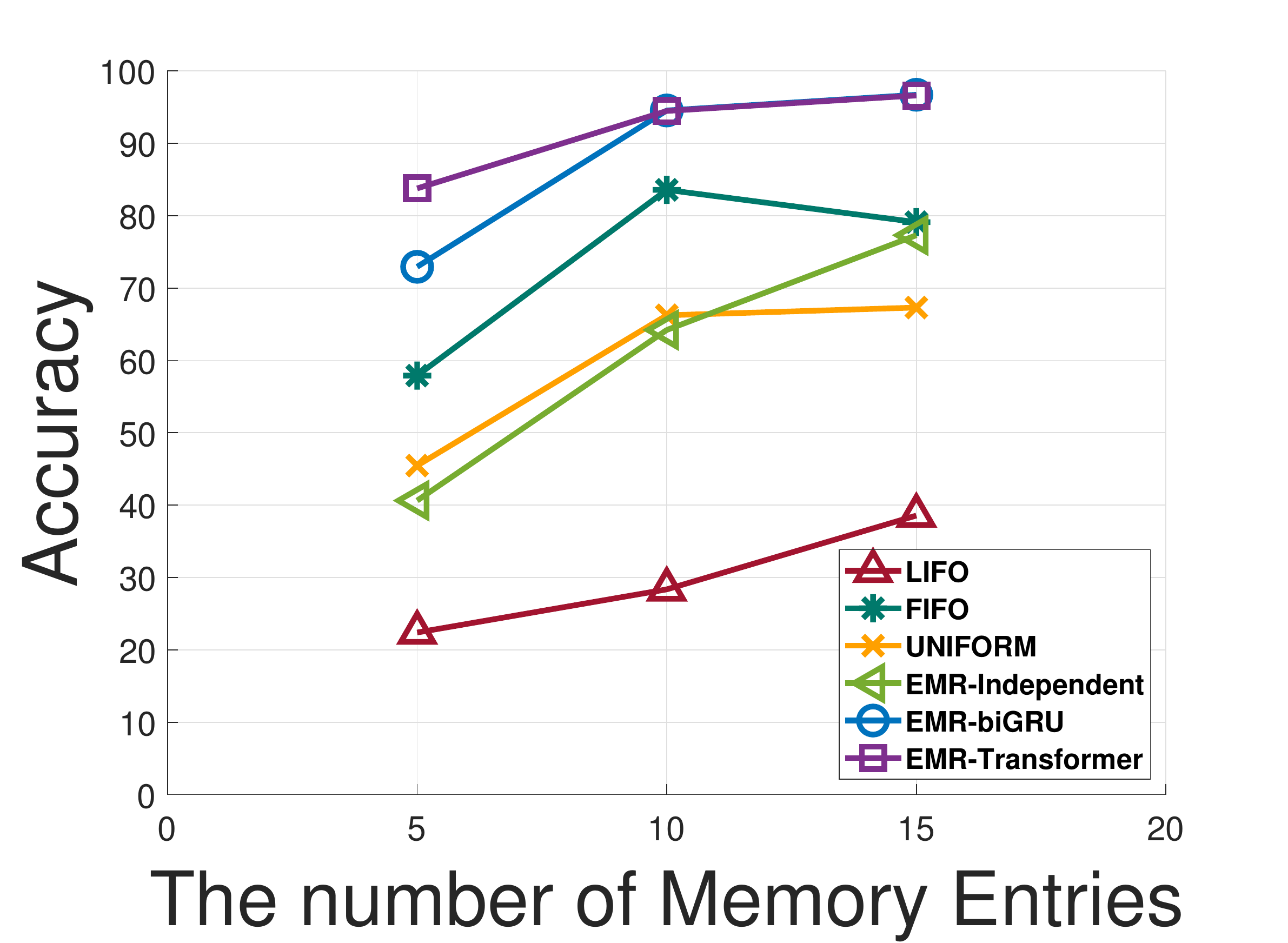} \\
		\caption{\small Original} \label{fig:babi/babi_task_2_acc}
	\end{subfigure}
	\vspace{-0.15cm}
	\begin{subfigure}[b]{0.23\textwidth}
		\centering
		\includegraphics[width=3.7cm]{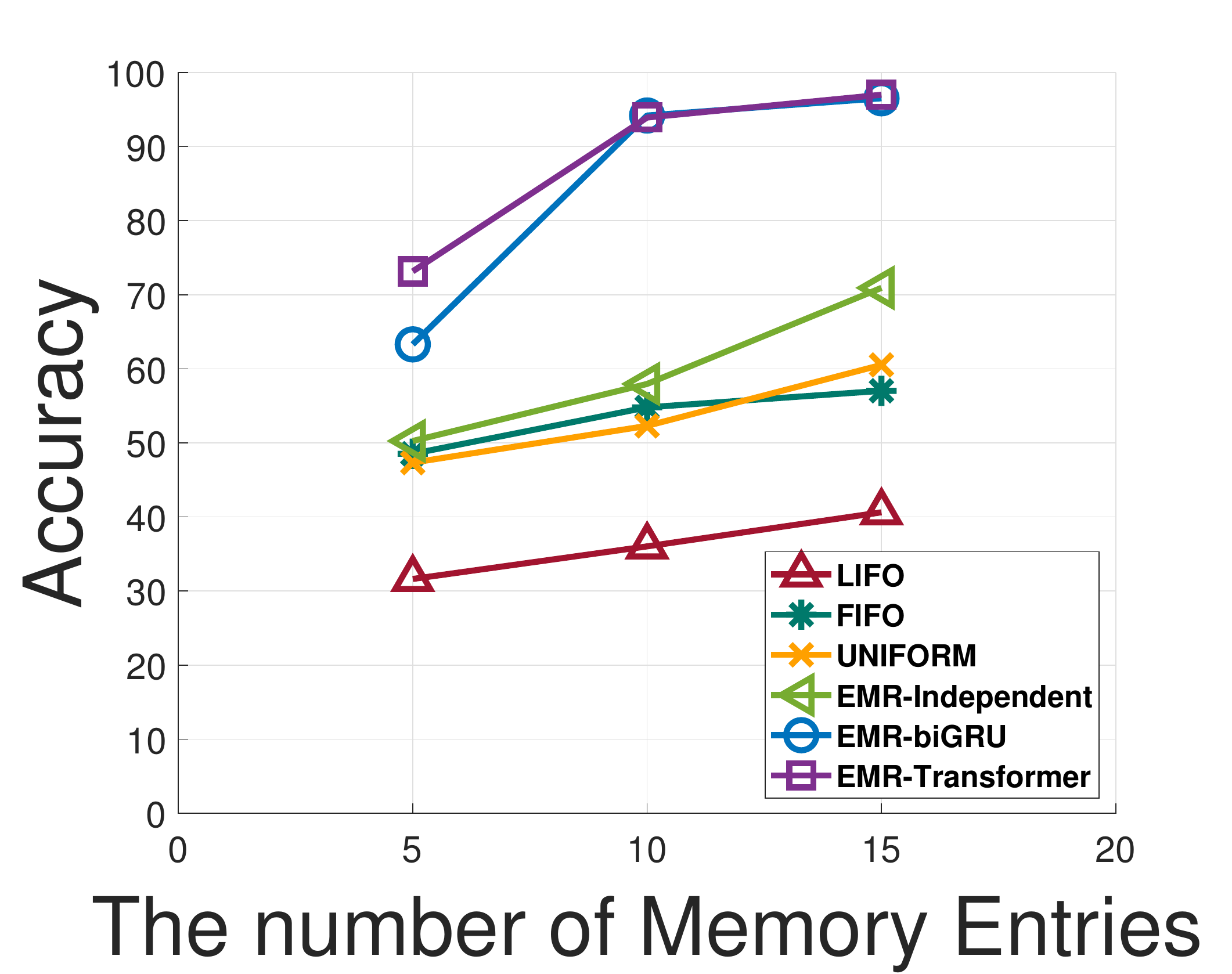} \\
		\caption{\small Noisy} \label{fig:babi/babi_task_22_acc}
	\end{subfigure}
	\caption{\small QA performance of baseline models and EMR-variants. The reported results are averages over 3 runs.}
	\label{fig:babi_result}
	\vspace{-0.30cm}
\end{figure}

\paragraph{Results and Analysis}
In Figure~\ref{fig:babi_result}, we report the experiment results for Original and Noisy tasks. Both our model (EMR-biGRU and EMR-Transformer) outperform the baselines, especially with higher gain in the case of Noisy dataset. EMR-independent, which does not consider relative importance among the memory entries, performs worse or simlar to FIFO baseline. The results suggest that our methods are able to retain the supporting facts even with small number of memory entries. For further analysis for the experiments on the bAbI dataset, please \textbf{see supplementary file}.


\subsection{TriviaQA \label{exp_trivia}}
\begin{figure}[!t]
	\centering
	{\includegraphics[width=0.95\linewidth]{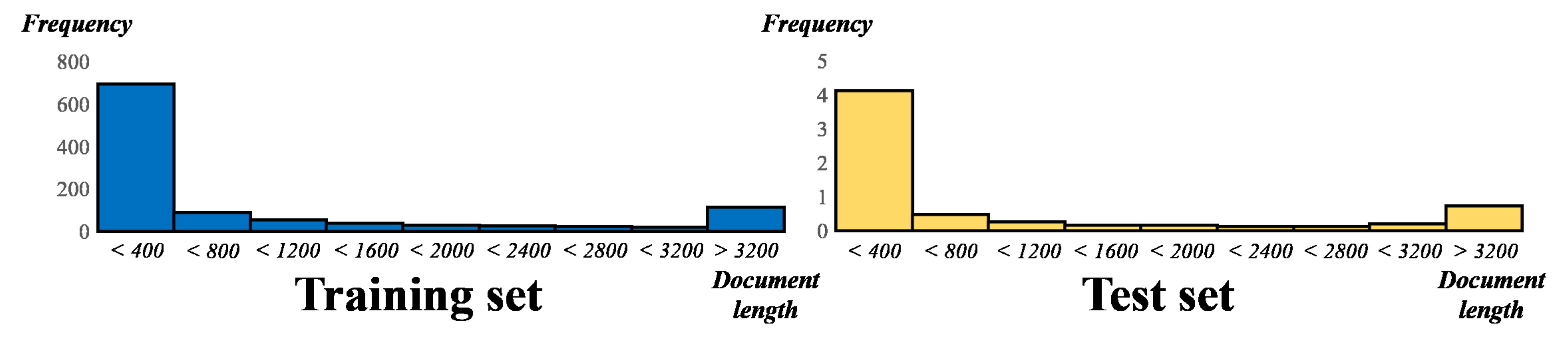}}
	\caption{\small The histogram of number of answers for each document length for TriviaQA dataset.}
	\label{fig:stats}
	\centering
\end{figure}
\paragraph{Dataset}
TriviaQA \cite{TriviaQA} is a realistic text-based question answering dataset which includes 950K question-answer pairs from 662K documents collected from Wikipedia and the web. This dataset is more challenging than standard QA benchmark datasets such as Stanford Question Answering Dataset (SQuAD) \cite{SQuAD}, as the answers for a question may not be directly obtained by span prediction and the context is very long (Figure~\ref{fig:trivia/example}). Since conventional QA models \cite{BiDAF, MemoReader, QANet, BERT} are span prediction models, on TriviaQA they only train on QA pairs whose answers can be found in the given context. In such a setting, TriviaQA becomes highly biased, where the answers are mostly spanned in the earlier part of the document (Figure~\ref{fig:stats}). We evaluate our work only on the Wikipedia domain since most previous work report similar results on both domains. While TriviaQA dataset consists of both human-verified and machine-generated QA subsets, we use the human-verified subset only since the machine-generated QA pairs are unreliable. We use the validation set for test since the test set does not contain labels.

\begin{figure*}[!ht]
	\centering
	\includegraphics[width=0.95\linewidth]{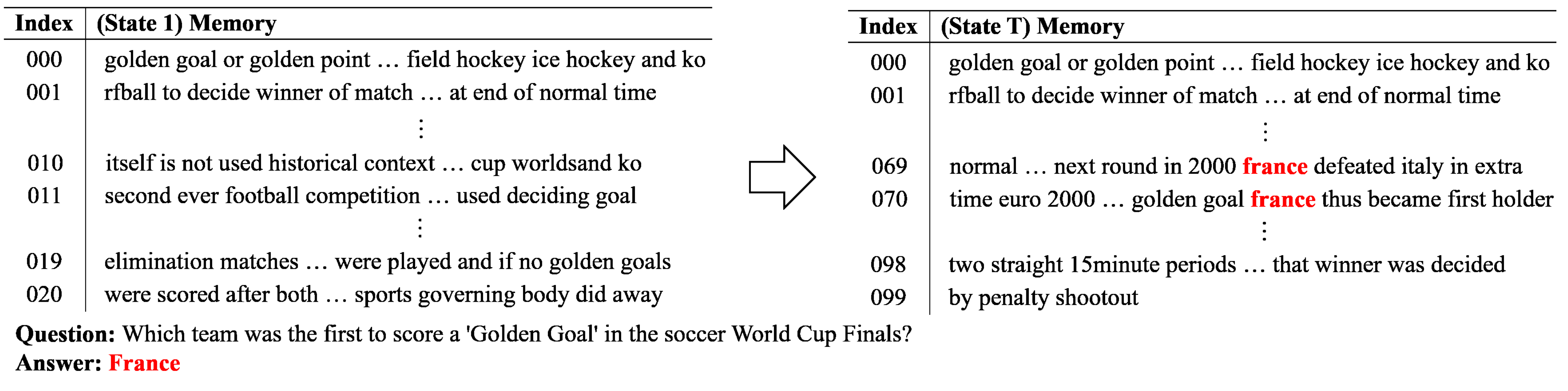} \\
	\vspace{-0.2cm}
	\caption{\small An example of how our model operates in order to solve the memory retention problem. Episodic Memory Reader (EMR) sequentially reads the sentences one by one while replacing least important memories. State 1 and State T represent memory entries in the in the initial state and the last state, respectively. Our EMR retrains the sentences which contain the word `France' (in bold red fonts) in order to answer the given question.}
	\label{fig:trivia_example}
	\vspace{-0.4cm}
\end{figure*}

\paragraph{Experiment Details}
We employ the pre-trained model from Deep Bidirectional Transformers (BERT) \cite{BERT}, which is the current state-of-the-art model for SQuAD challenge, that trains several Transformers in \citet{Self-Attention} for pretraining tasks for predicting the indices of the exact location of an answer. We embed $20$ words into each memory cell using a GRU and set the number of cells to $20$, thus the memory can hold $400$ words at maximum. This is a reasonable restriction since BERT limits the maximum number of word tokens to $512$, including both the context and the query.

\paragraph{Results and Analysis}
We report the performance of our model on the TriviaQA using both ExactMatch and F1-score in Table \ref{tab:trivia/result}. We see that EMR models which consider the relative importance between the memory entries (EMR-biGRU and EMR-Transformer) outperform both the rule-based baselines and EMR-Independent. One interesting observation is that LIFO performs quite well unlike the other rule-based scheduling policies, but this is due to the dataset bias (See Figure~\ref{fig:stats}) where most answers are spanned in earlier part of the documents. To further see whether this improvement is from its ability to remember important context, we examine the sentences that remain in the memory after EMR finishes reading all the sentences in Figure~\ref{fig:trivia_example}. We see that EMR remembered the sentences that contain key words that is required to answer the future question.

\begin{table}[t]
	\centering
	\caption{\small Q\&A accuracy on the TriviaQA dataset. model.} \label{tab:trivia/result}
	\begin{tabular}{ccc}
		Model           & ExactMatch         & F1        \\ \hline \hline
		FIFO            & 24.53      & 27.22     \\
		Uniform         & 28.30      & 34.39     \\ 
		LIFO            & 46.23      & 50.10     \\
		EMR-Independent & 38.05      & 41.15     \\	
		\hline
		EMR-biGRU       & \bf 52.20  & \bf 57.57 \\	
		EMR-Transformer & 48.43      & 53.81     \\
		\hline
	\end{tabular}
\end{table}



\subsection{TVQA \label{exp_tvqa}}
\begin{figure*}[!ht]
	\centering
	\begin{tabular}{@{}c@{}}
		\includegraphics[width=14.5cm]{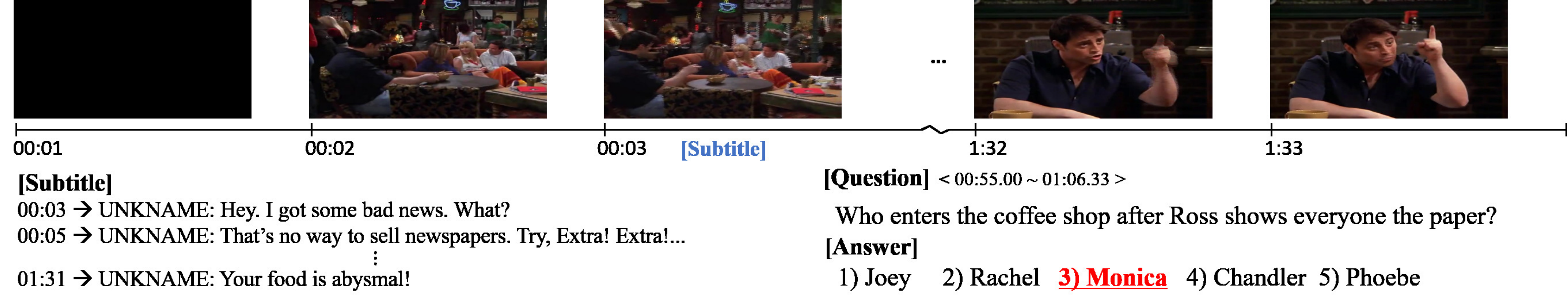} \\
		\vspace{-0.7cm}
	\end{tabular}
	\begin{tabular}{@{}c@{}}
		\includegraphics[width=14.5cm]{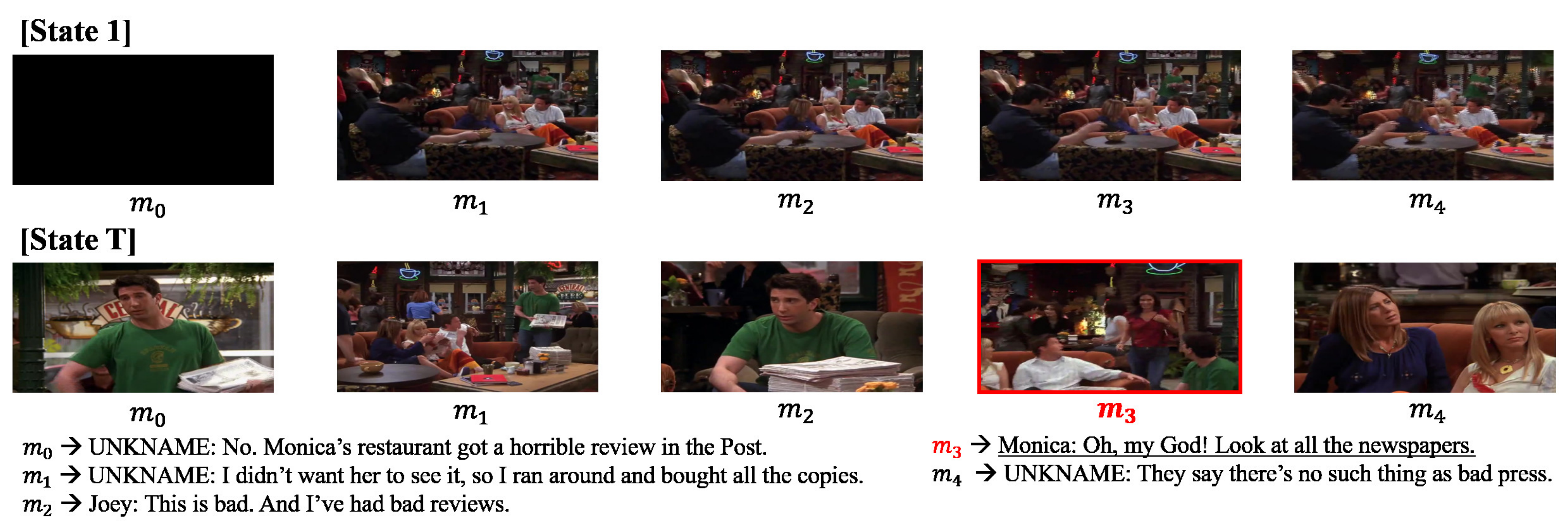} \\
	\end{tabular}
	\vspace{-0.3cm}
	\caption{\small An example of TVQA dataset and a visualization of how our model operates. The answer in red is the ground truth and the answer with underline is the predicted answer from our QA model. At state 1, since the memory is empty, our model encodes every video frame into the memory until the memory becomes full. At state T, after reading in all the video frames, it retains the most informative frames to answer the question. Note that this retention of the important frames is done without knowing the question in advance.}
	\centering
	\label{fig:tvqa/figure}
	\vspace{-0.5cm}
\end{figure*}  

\paragraph{Dataset}
TVQA \cite{TVQA} is a localized, compositional video question-answering dataset that contains 153K question-answer pairs from 22K clips spanning over 460 hours of video. The questions are multiple choice questions on the video contents, where the task is to find a single correct answer out of five candidate answers. The questions can be answered by examining the annotated clip segments, which spans around $30$ frames per clip (See Figure~\ref{fig:tvqa/figure}). The average number of frames for each clip is $229$. In addition to the video frames, the dataset also provides subtitles for each video frame. Thus solving the questions requires compositional reasoning capability over both a large number of images and texts. 

\begin{figure}[t]
	\centering
	\includegraphics[width=6.5cm]{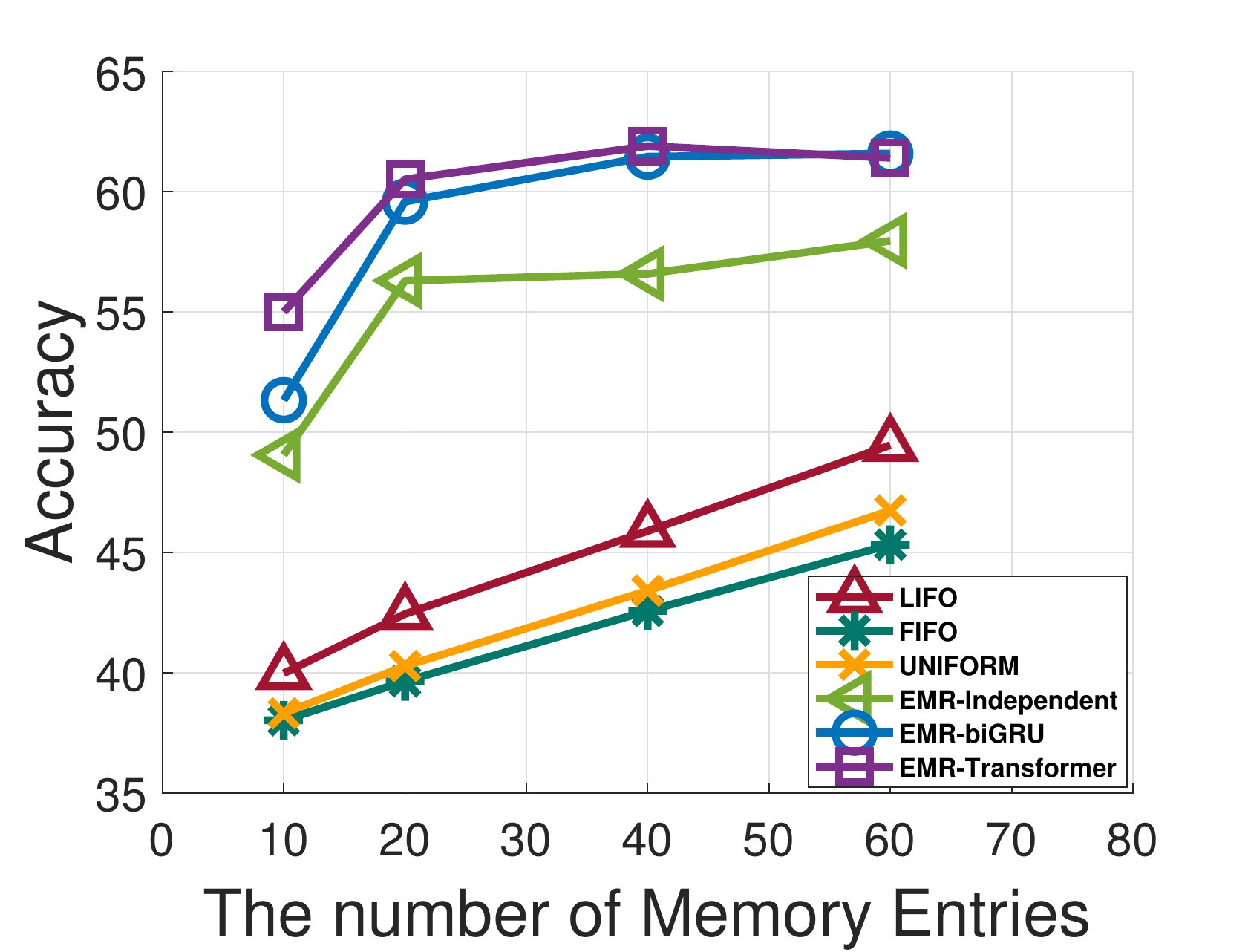}	
	\vspace{-0.10in}
	\caption{\small QA accuracy of various memory scheduling policies on the TVQA dataset, reported as a function of the number of memory entries.}
	\label{tab:tvqa/result}
	\vspace{-0.6cm}
\end{figure}

\paragraph{Experiment Details}
As for the QA module, we use Multi-stream model for Multi-Modal Video QA, which is the attention-based baseline model provided in~\cite{TVQA}. For efficient training, we use features extracted from a ResNet-101 pretrained on the ImageNet dataset. For embedding subtitles and question-answering pairs, we use GloVe~\cite{Glove}. For training, we restrict the number of memory entries for our episodic reader as $20$, where each memory entry contains the encoding of a video frame and the subtitle associated with the frame, where the former is encoded using CNN and the latter using GRU. We train our model and the baseline models using the ADAM optimizer \cite{Adam}, with the initial learning rate of $0.0001$. Unlike from the experiments on TriviaQA, we use REINFORCE \cite{REINFORCE} to train the policy. This is because TVQA is composed of consecutive image frames captured within a short time interval, which tend to contain redundant information. Thus the value network of the actor-critic model fails to estimate good value of the given state since deleting a good frame will not result in the loss of QA accuracy. Thus we compute the reward $\mathcal{R}^{(t)}$ as the accuracy difference between at time step $t$ and $t-1$ then use only the policy with non-episodic REINFORCE for training. With this method, if the task fails to solve the question after deleting certain frame, the frame is considered as important, and unimportant otherwise.


\paragraph{Results and Analysis}
We report the accuracy on TVQA as a function of memory size in Figure \ref{tab:tvqa/result}. We observe that EMR variants significantly outperform all baselines, including EMR-Independent. We also observe that the models perform well even when the size of the memory is increased to as large as $60$, which was never encountered during the training stage where the number of memory entries was fixed as $20$. When the size of memory is small, the gap between different models are larger, with EMR-Transformer obtaining the best accuracy, which may be due to its ability to capture global relative importance of each memory entry. However, the gap between EMR-Transformer and EMR-biGRU diminishes as the size of memory increases, since then the size of the memory becomes large enough to contain all the frames necessary to answer the question. 

As qualitative analysis, we further examine which frames and subtitles were preserved in the external memory after the model has read through the entire sequence in Figure~\ref{fig:tvqa/figure}. To answer the question for this example, the model should consider the relationship between two frames, where the first frame describes Ross showing the paper to others, and the second frame describes Monica entering the coffee shop. We see that our model kept both frames, although it did not know what the question will be. 

\section{Conclusion}
We proposed a novel problem of question answering from streaming data, where the model needs to answer a question that is given after reading through unlimited amount of context (e.g. documents, videos) that cannot fit into the system memory. To handle this problem, we proposed \emph{Episodic Memory Reader (EMR)}, which is basically a memory-augmented network with RL-based memory-scheduler, that learns the relative importance among memory entries and replaces the entries with the lowest importance to maximize the QA performance for future tasks. We validated EMR on three QA datasets against rule-based memory scheduling as well as an RL-baseline that does not model relative importances among memory entries, which it significantly outperforms. Further qualitative analysis of memory contents after learning confirms that such good performance comes from its ability to retain important instances for future QA tasks.

\section*{Acknowledgements}
This work was supported by Clova, NAVER Corp. and Institute for Information \& communications Technology Promotion(IITP) grant funded by the Korea government(MSIT) (No.2016-0-00563, Research on Adaptive Machine Learning Technology Development for Intelligent Autonomous Digital Companion).

\bibliography{acl2019}
\bibliographystyle{acl_natbib}

\appendix
	\appendix
		\begin{figure*}[!ht]
		\centering
		\begin{subfigure}[]{0.49\textwidth}
			\centering
			\includegraphics[width=4.2cm]{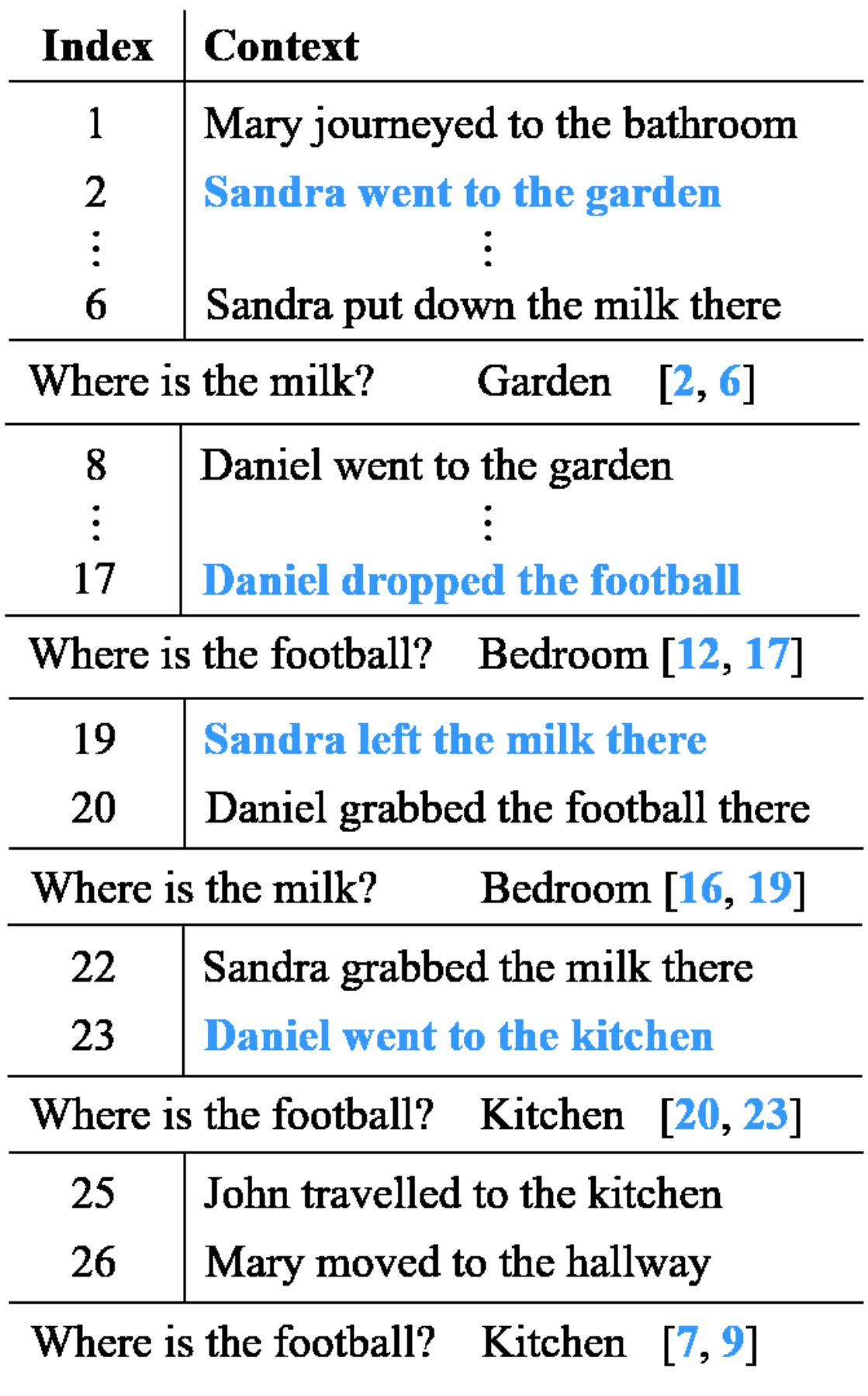} \\
			\caption{\small Original} \label{fig:babi_original}
		\end{subfigure}
		\begin{subfigure}[]{0.49\textwidth}
			\centering
			\includegraphics[width=4.2cm]{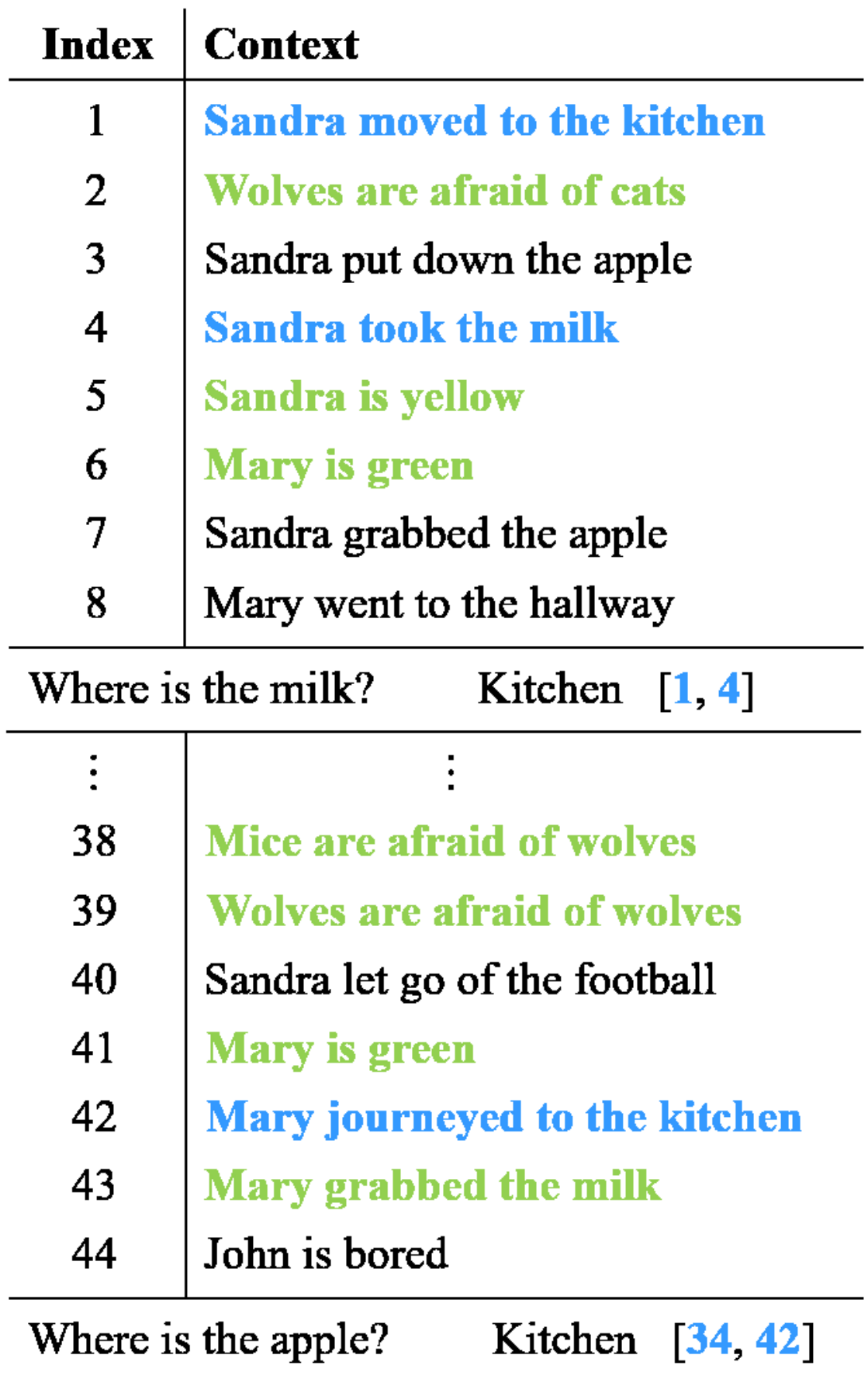} \\
			\caption{\small Noisy} \label{fig:babi_noisy}
		\end{subfigure}
		\caption{Example of (a) Original task and (b) Noisy task. Sentences in green are noise sentences and ones in blue are supporting facts of each question.}
		\label{fig:babi_dataset}
		\vspace{-0.5cm}
	\end{figure*}
	
	\section{bAbI dataset}
	We provide examples of the Original and Noisy datasets, as well as visualization of the memorized examples to show what our EMR models have remembered, for bAbI \cite{bAbI} dataset.
	
	\paragraph{Dataset}
	We visualize an example for \textbf{Original} and \textbf{Noisy} tasks in Figure~\ref{fig:babi_dataset}.

\begin{figure*}[!ht]
		\centering
		\begin{subfigure}[]{0.49\textwidth}
			\centering
			\includegraphics[width=6.0cm]{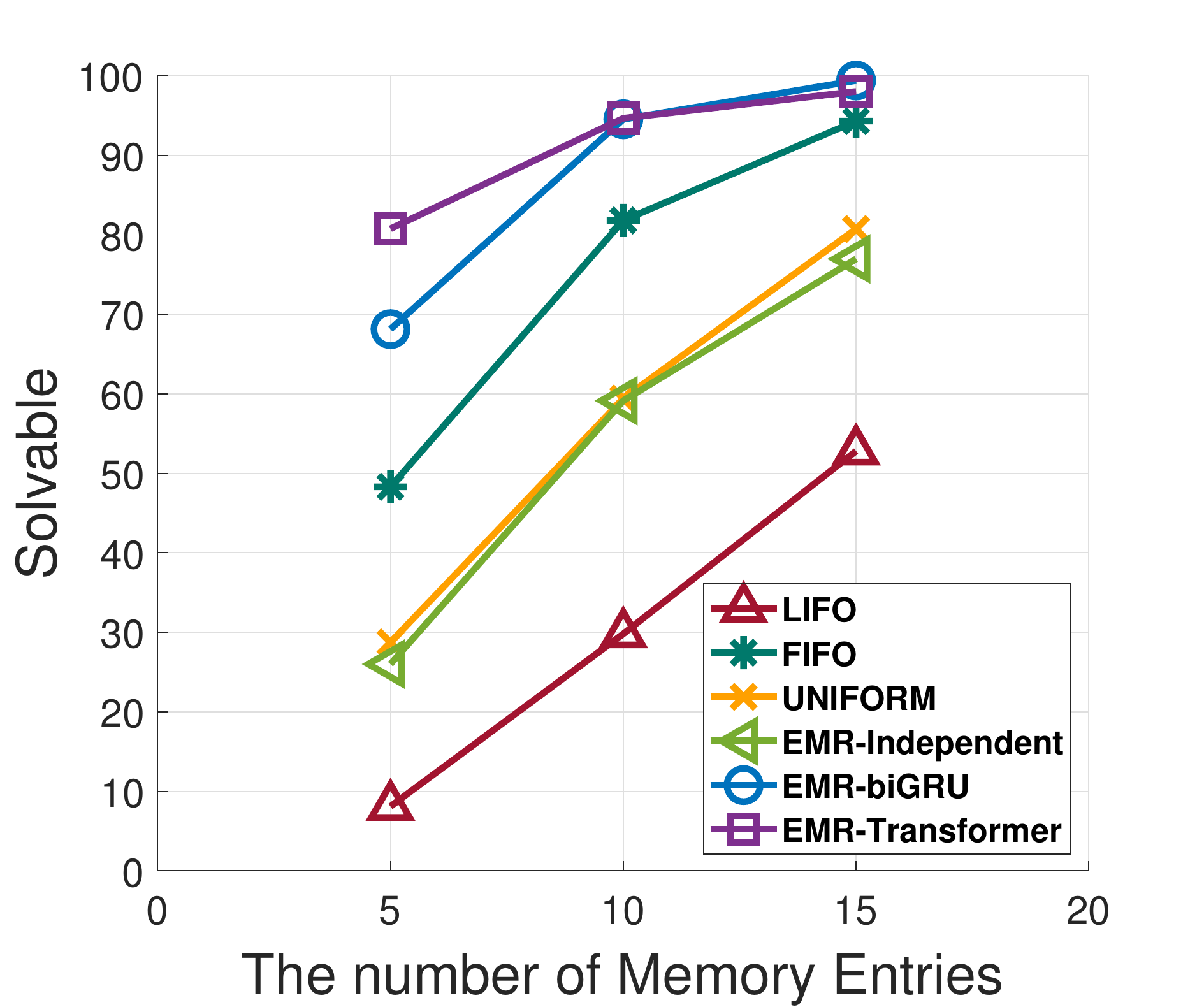} \\
			\caption{\small Original (Solvable)} \label{fig:babi_task_2_sup}
		\end{subfigure}
		\begin{subfigure}[]{0.49\textwidth}
			\centering
			\includegraphics[width=6.0cm]{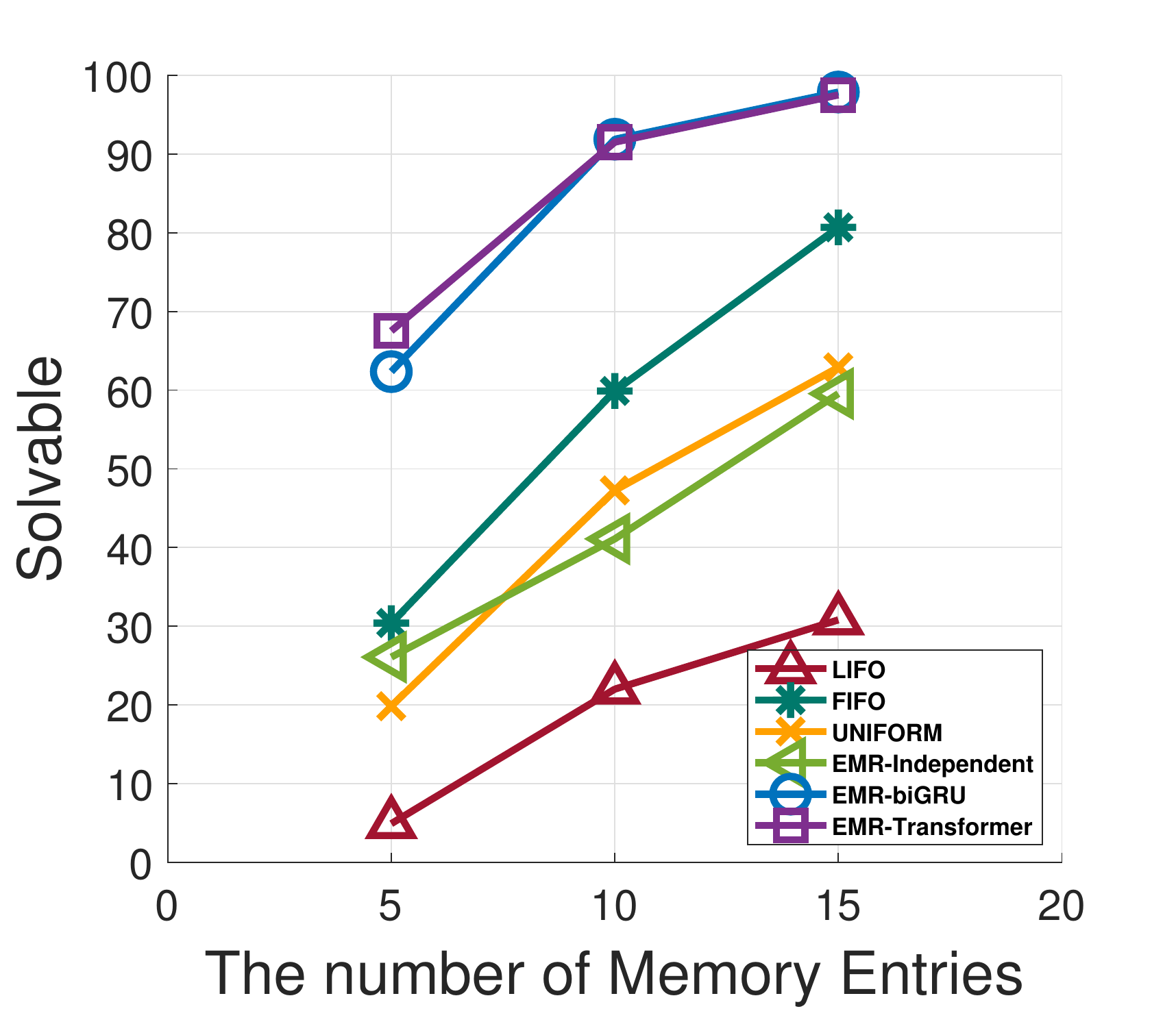} \\
			\caption{\small Noisy (Solvable)} \label{fig:babi_task_22_sup}
		\end{subfigure}
		\caption{The results for our model (EMR-biGRU and EMR-Transformer) and the baselines. The reported results are averages over 3 runs. The Solvable represents an accuracy that when the model encounters a question, it contains supporting facts in the memory to solve the question.}
		\label{fig:babi_result_sup}
	\end{figure*}

	
	\paragraph{Results and Analysis}
	As shown in Figure~\ref{fig:babi_result_sup}, we further report the performance of the baseline models and our EMR variants, on how many supporting facts they retrain in the memory (denoted as solvable), by considering the QA performance with a perfect QA model. We observe that both EMR variants, EMR-Independent and EMR-Transformer, significantly outperform rule-based memory scheduling agents as well as EMR-Independent.
	
	\begin{figure*}[!ht]
	\centering
	\begin{subfigure}[]{0.49\textwidth}
		\centering
		\includegraphics[width=4.2cm]{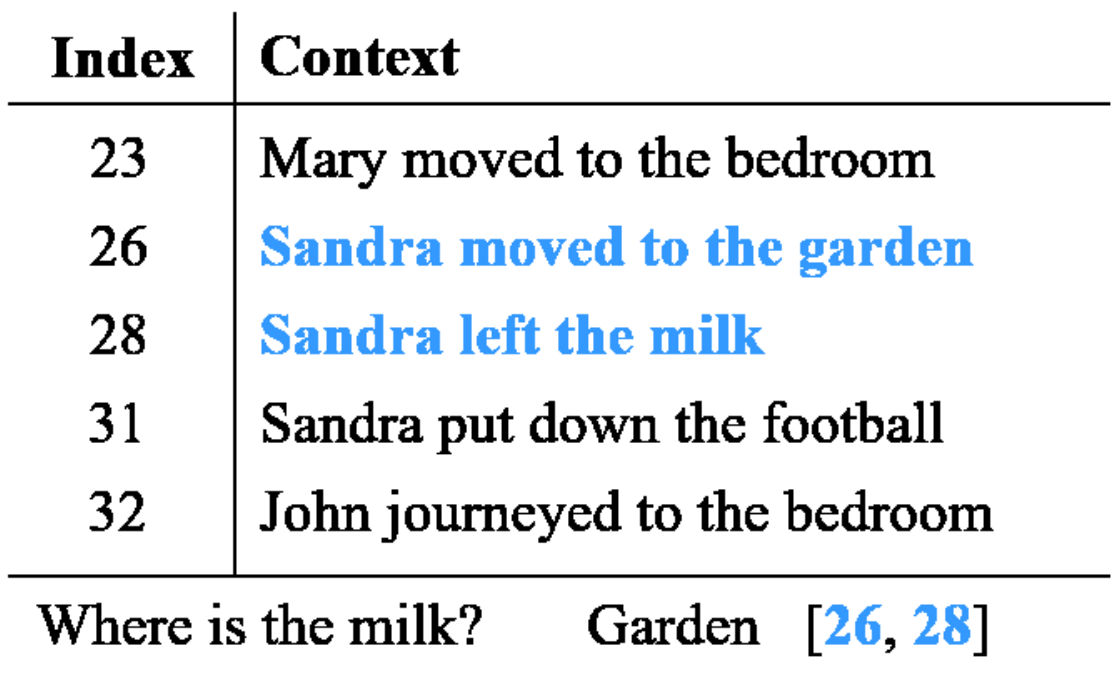} \\
		\caption{\small EMR-biGRU (Original)} \label{fig:bigru_ori}
		\vspace{0.5cm}
	\end{subfigure}
	\begin{subfigure}[]{0.49\textwidth}
		\centering
		\includegraphics[width=4.2cm]{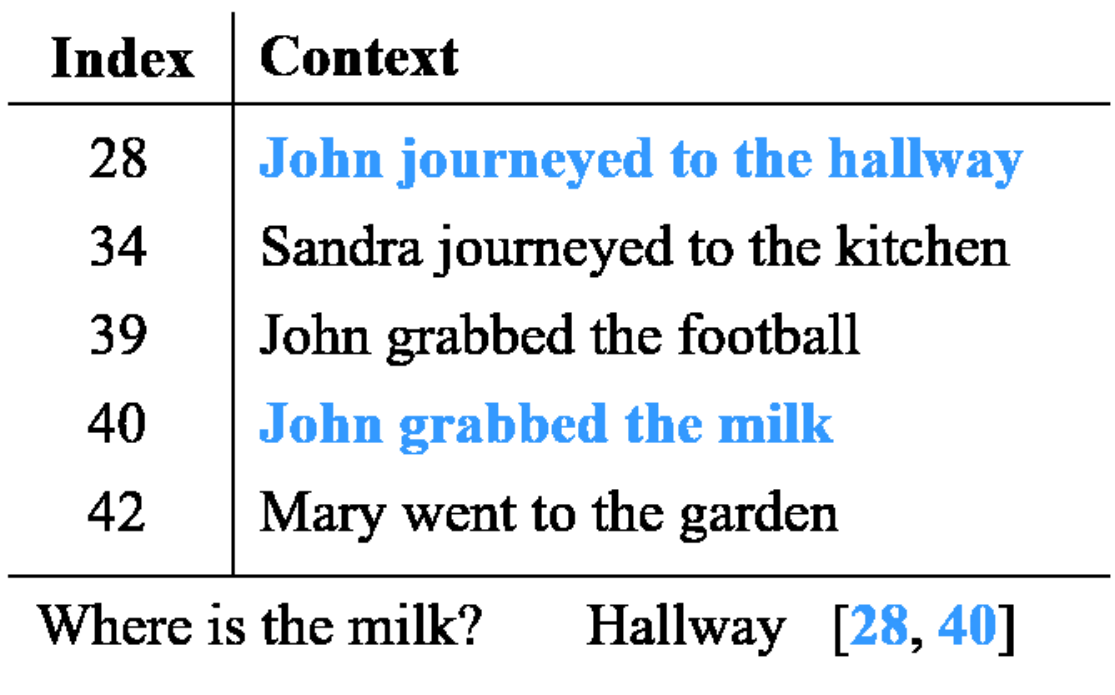} \\
		\caption{\small EMR-biGRU (Noisy)} \label{fig:bigru_noisy}
		\vspace{0.5cm}
	\end{subfigure}
	\begin{subfigure}[]{0.49\textwidth}
		\centering
		\includegraphics[width=4.2cm]{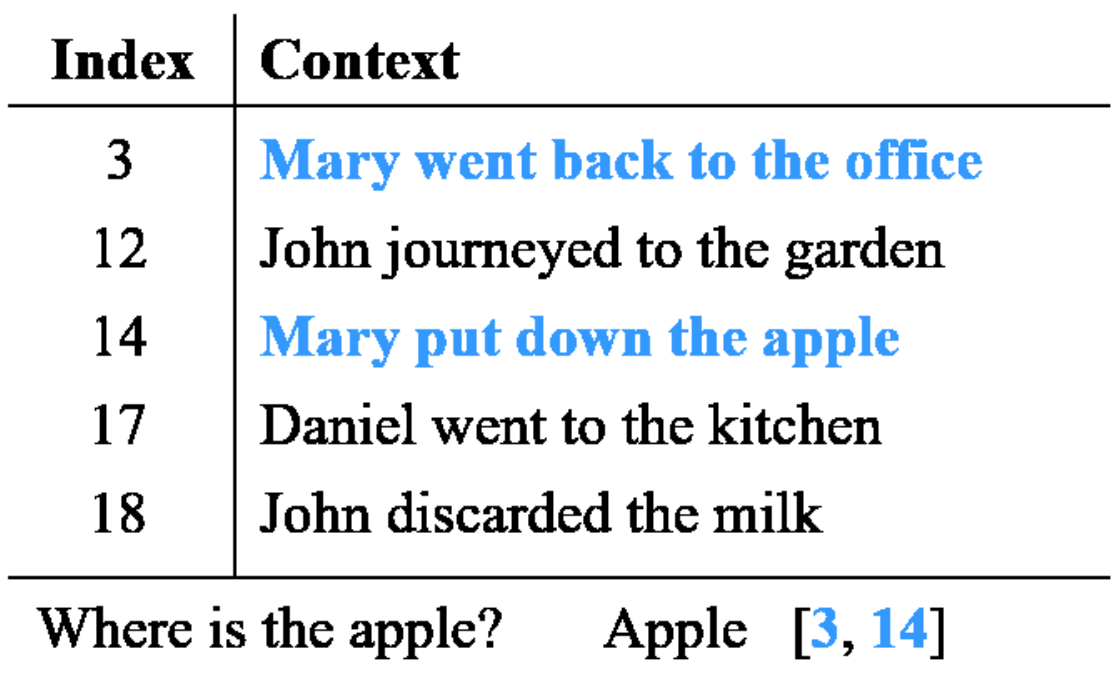} \\
		\caption{\small EMR-Transformer (Original)} \label{fig:trans_ori}
	\end{subfigure}
	\begin{subfigure}[]{0.49\textwidth}
		\centering
		\includegraphics[width=4.2cm]{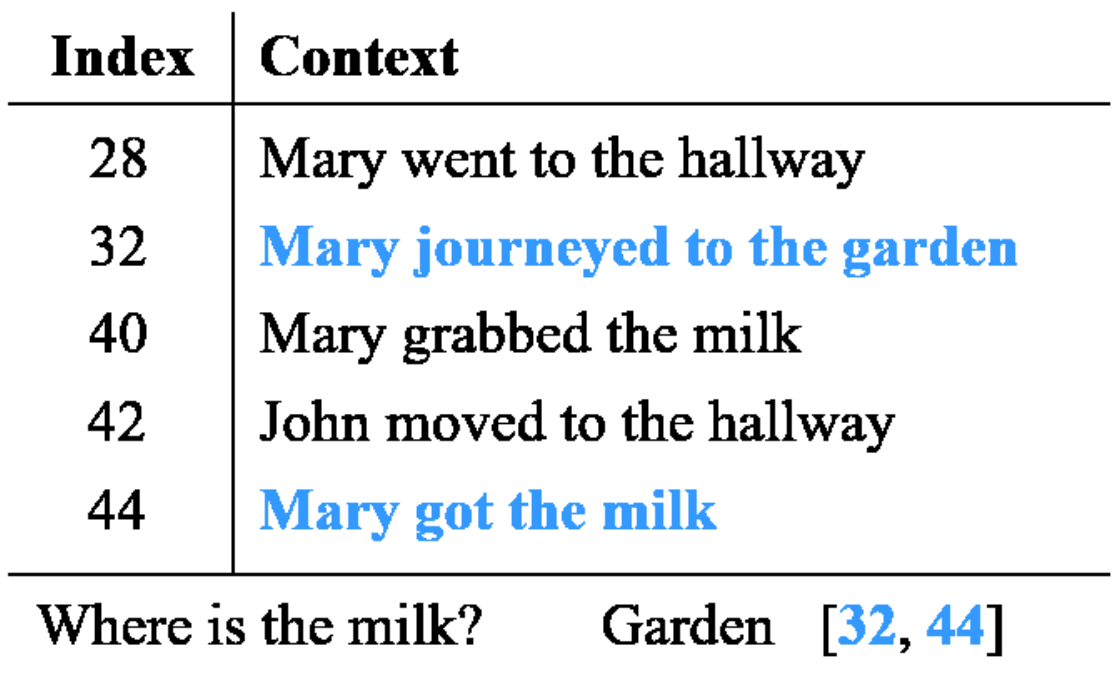} \\
		\caption{\small EMR-Transformer (Noisy)} \label{fig:trans_noisy}
	\end{subfigure}
	\caption{Example of Original and Noisy task for EMR-biGRU and EMR-Transformer. Sentences in blue are supporting facts of each question. The Index on the figure represents the order of sentences in the context.} \label{fig:babi_examples}
	\end{figure*}
	
	\section{TriviaQA}
	We provide more experiment details and additional examples for analyzing what our EMR models have remembered, for TriviaQA \cite{TriviaQA}.
	
	\paragraph{Dataset}
	The objective of our model is to learn general importance in situations where not knowing the question from streaming data. In terms of scalability, our model is able to access sequentially a large amount of streaming data by replacing the most uninformative memory entry in the external memory. When comparing TriviaQA with a common question-answering dataset \citep{SQuAD, bAbI}, it is an appropriate dataset to prove the efficiency of our model since its average word number is approximately 3K which cannot be accessed using conventional models that predicts answer indices using a pointer network \citep{BiDAF, MemoReader, QANet}.
	
	To preprocess TriviaQA according to problem setting, we truncate all documents within 1200 words for a training set, in order to reduce the cost of training process. Unlike the training set, a test set takes all words in the documents. Although TriviaQA does not provide the answer indices in a document, we extract the documents that can be spanned to adopt Deep Bidirectional Transformers (BERT) \cite{BERT}, which is state-of-the-art reading comprehension model using a pointer network. Additionally, we made all letters lowercase and removed all special characters.

	\paragraph{Experiment Details}
	As described in the main paper, we employed the pre-trained BERT to solve TriviaQA. A more specific implementation is described here. We encode the current input $x^{(t)}$ to $m_{i}^{(t)}$ using the BERT encoding layer and a bi-directional GRU whose output size is 768 and 128, respectively. The reason for using it is to convert the words into a sentence. By doing this, it can make accessing a possible chunk of words and computation cost is reduced. We utilize $m_{i}^{(t)}$ to output relative importance between the memory entries \{$m_{1}^{(t)}, ..., m_{i}^{(t)}$\}, where ${i}$ indicates an address in the memory entry, as described in the main paper. In addition to using the pre-trained BERT, we finetune it with truncated documents (Up to 400 words) in the same way as LIFO (Last-In-First-Out) since hoping our model focuses on learning what to remember in the external memory and generalizes well even watching limited contents in the documents. We train our model and the baseline models using ADAM optimizer \cite{Adam}, with the initial learning rate of 1e-5 and dropout probability of 0.1 for 1M steps. For A3C \cite{A3C}, we set the discount factor to 0.1 and entropy regularization to 0.01 for all experiments.
	
\begin{figure}[t]
		\centering
		\includegraphics[width=6.5cm]{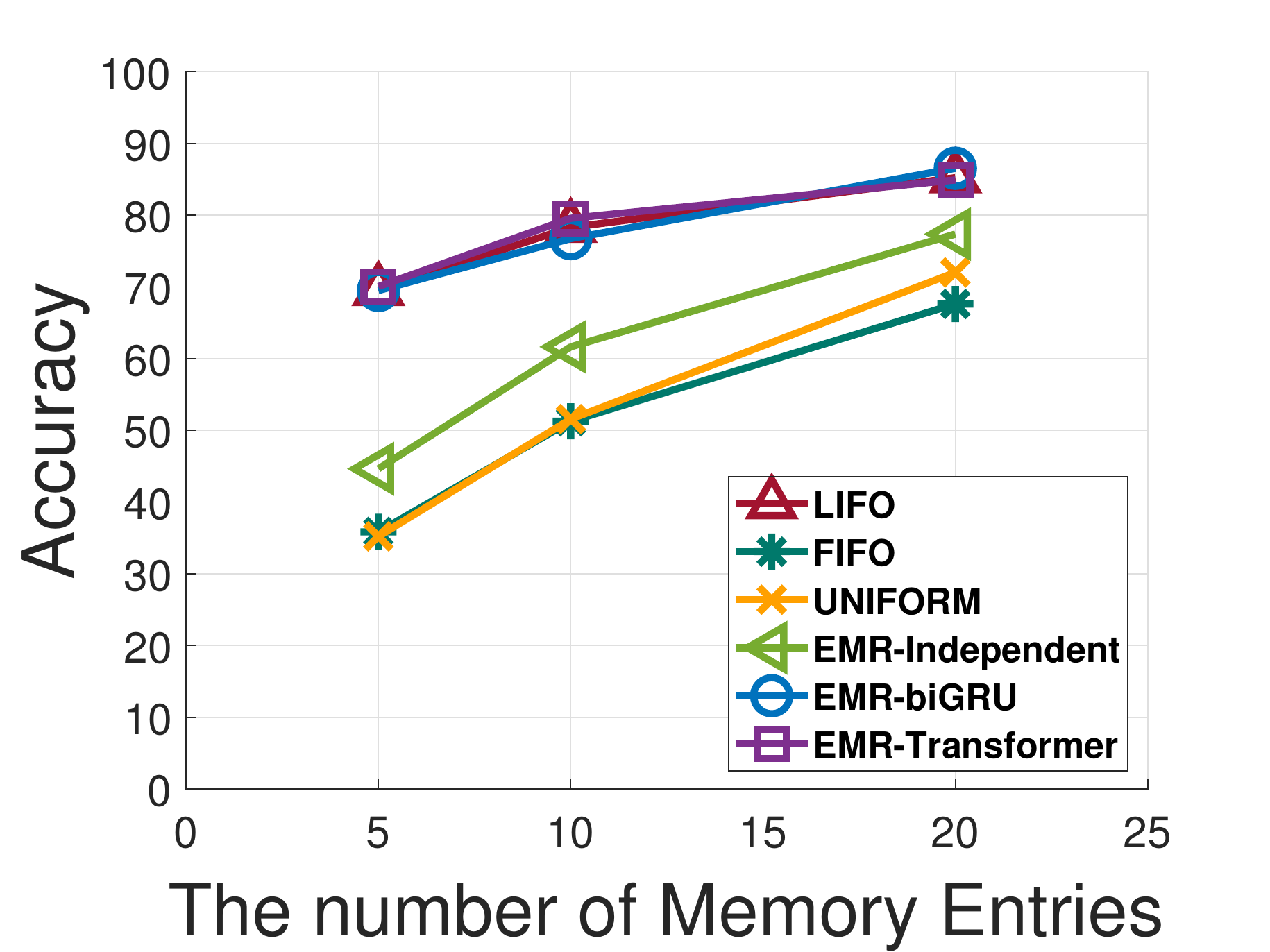}	
		\vspace{-0.10in}
		\caption{\small Oracle score }
		\label{tab:trivia_oracle}
		\vspace{-0.6cm}
	\end{figure}

	\paragraph{Results and Analysis}
	As shown in Figure~\ref{tab:trivia_oracle}, we report the score of each method using a perfect QA model, to see how many of the important facts are remembered by each method. We see that on TriviaQA dataset, LIFO contains similar amount of words as EMR-biGRU and EMR-Transformer. This is mostly due to the dataset bias, where most of the answers are found in earlier parts of the documents (Figure 6). However, our models outperform LIFO in QA task, since it observed more sentences during training which help the QA model to perform better, compared to LIFO that observed less number of training examples during training due to Last-In-First-Out policy that discards all words that come after the memory is filled.
	
	\section{TVQA}
	We provide more experimental details and examples to show what our EMR models have remembered for the TVQA dataset. Each frame illustrated in the figure are the frames in the external memory at the last time step. The stars with different colors denote the supporting frames for different questions.

	\paragraph{Experiment Details}
	As described in the main paper, we use the Multi-stream model for Multi-Modal Video QA, which is suggested in \citet{TVQA}. We also pretrain the QA model for a delicate check of the performance of our EMR model. We use only the annotated frame when training the QA model. Since we use only the subtitle and frame image feature as input, we pretrain the QA model until reaching the reported performance of S+V model with the annotated time stamp in \citet{TVQA}.
	
	Below is the detailed implementation of our model EMR. Since we have two kinds of input $x^{(t)}$ in TVQA, we need to blend them to one memory feature to be fitted to our model. In the case of subtitle input, we use GloVe \cite{Glove} to embed words to 300-size vectors. Then, we use bi-directional GRU to make the sentence 128-size vector from word vectors. In the case of video frame input, we use 2048-size feature vectors extracted from a ResNet-101 pretrained on the ImageNet dataset. Then, we compress video frame vectors to 128-size vector using a linear layer. Then, we add two 128-size feature vectors from the subtitle and the video frame to make 128-size of memory feature vector $m_{i}^{(t)}$. Other details including optimizer and reinforcement learning setting are described in the main paper.

	
	\newpage
	\begin{figure*}[h!]
		\centering
		\includegraphics[width=\linewidth]{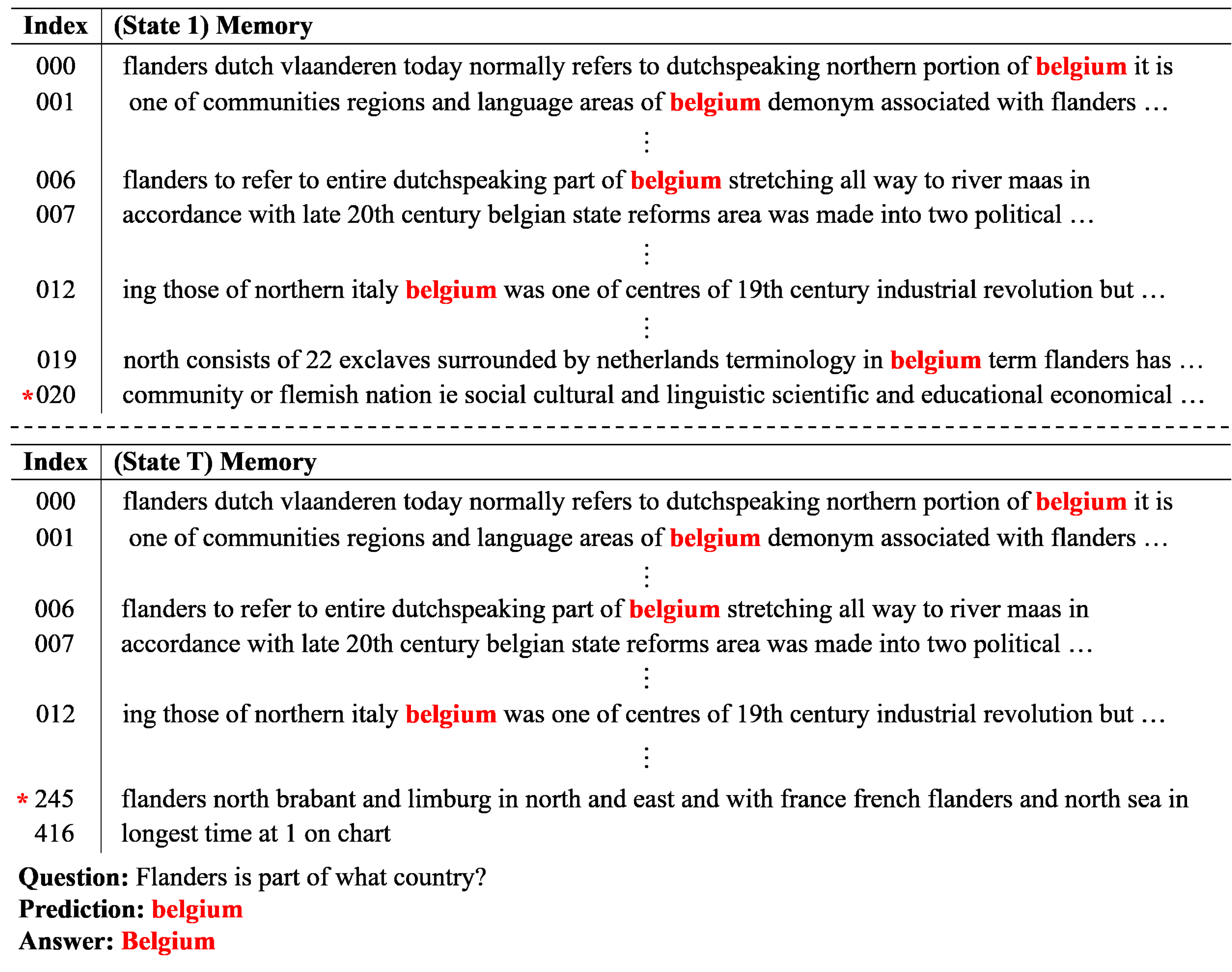} \\
		\caption{\small An example visualization of the memory. The answer word 'belgium' (Red / Thick) arrives at first timestep, and our model retains sentences at state T, which means after reading all the contexts. The star shape (*) indicates our model's selection which memory entry is deleted.}
		\centering
	\end{figure*}

	\begin{figure*}[h!]
		\centering
		\includegraphics[width=\linewidth]{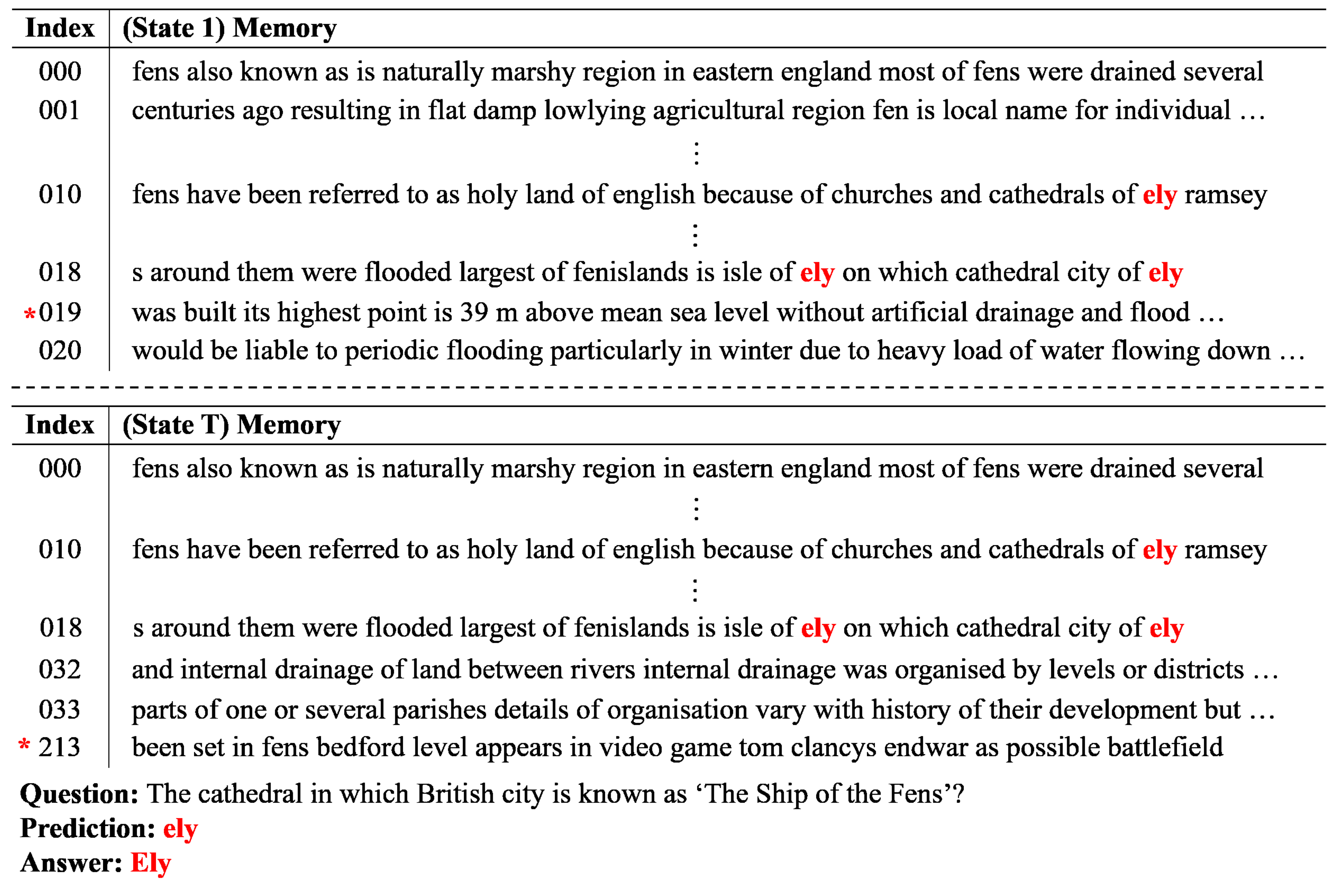} \\
		\caption{\small An example visualization of the memory. The answer word 'ely' (Red / Thick) arrives at first timestep, and our model retains it after reading in all the context sentences. The star shape (*) indicates our model's selection which memory entry is deleted.}
		\vspace{0.5cm}
		\includegraphics[width=\linewidth]{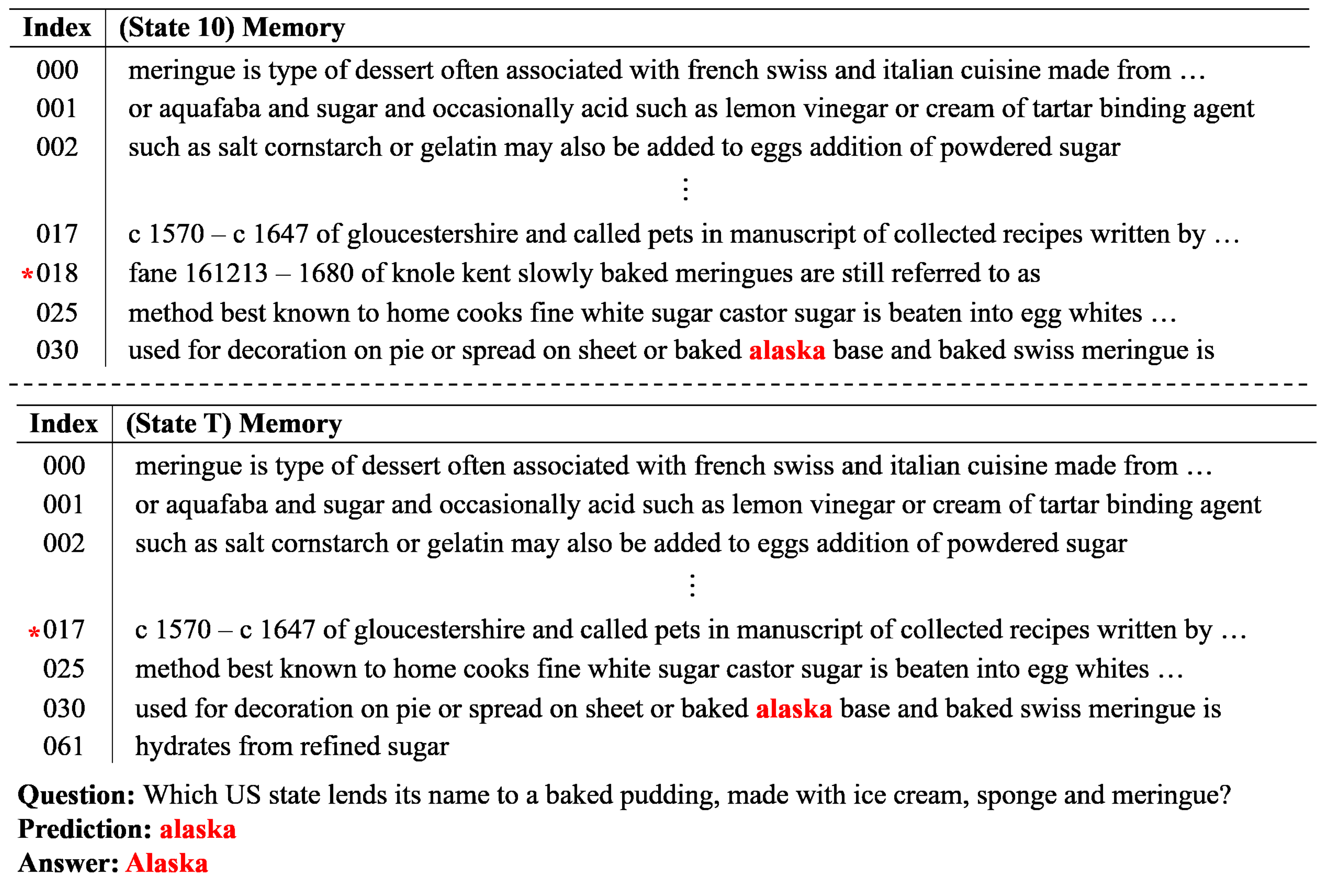} \\
		\caption{\small An example visualization of the memory. The answer word 'alaska' (Red / Thick) arrives at timestep 10, and our model retains it after reading in all the context sentences. The star shape (*) indicates our model's selection which memory entry is deleted.}
		\centering
	\end{figure*}
	
	\begin{figure*}[h!]
		\centering
		\begin{tabular}{@{}c@{}}
			\includegraphics[width=\linewidth]{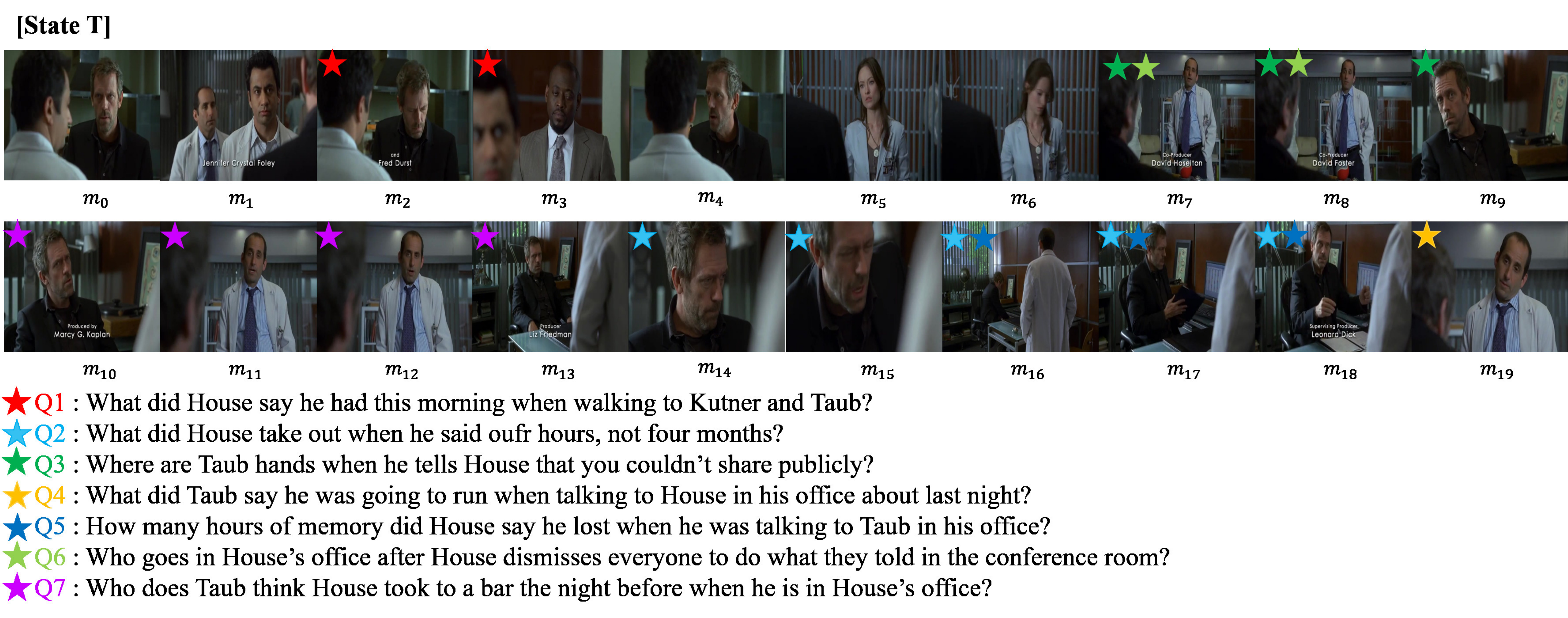} \\
		\end{tabular}
		\caption{\small An example of clip from drama 'House'. Each frame with star is corresponding to question with the star of same color.}
		\centering
	\end{figure*}
	
	\begin{figure*}[h!]
		\centering
		\begin{tabular}{@{}c@{}}
			\includegraphics[width=\linewidth]{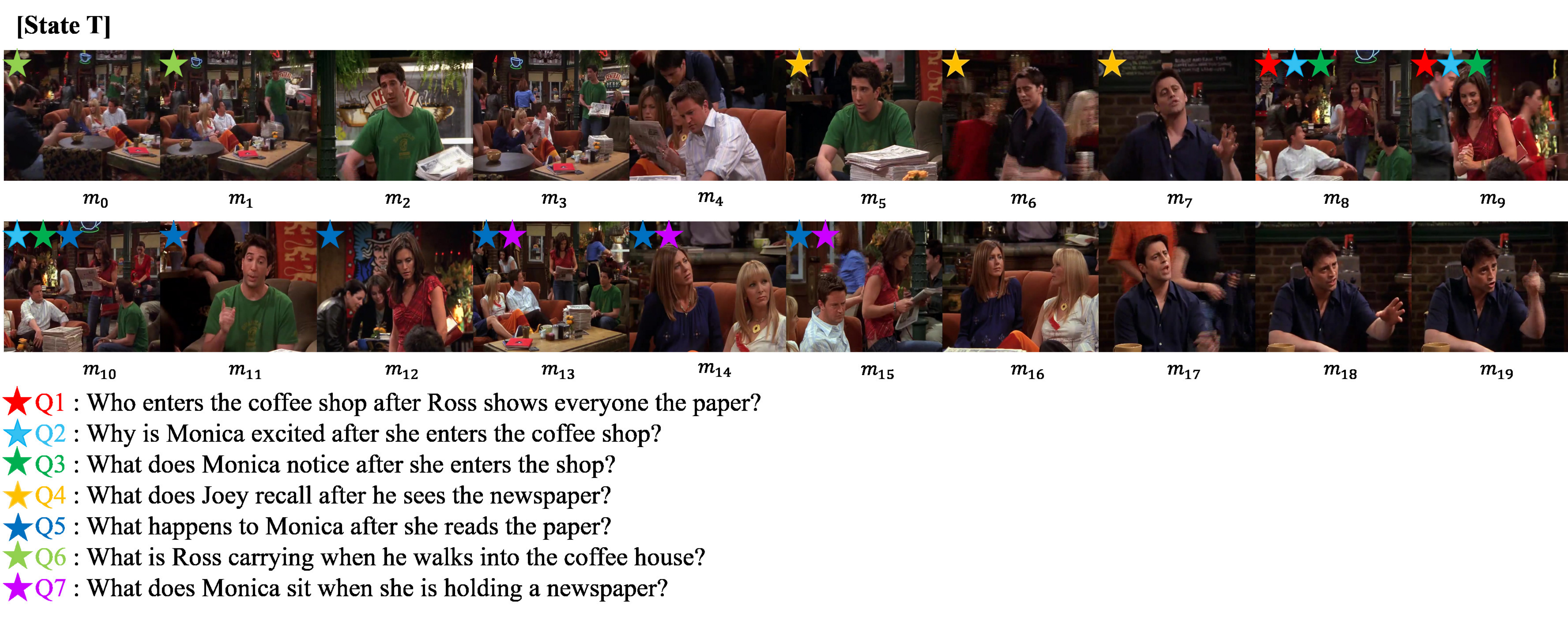} \\
		\end{tabular}
		\caption{\small An example of clip from drama 'Friends'. Each frame with star is corresponding to question with the star of same color. }
		\centering
	\end{figure*}
	
	\begin{figure*}[h!]
		\centering
		\begin{tabular}{@{}c@{}}
			\includegraphics[width=\linewidth]{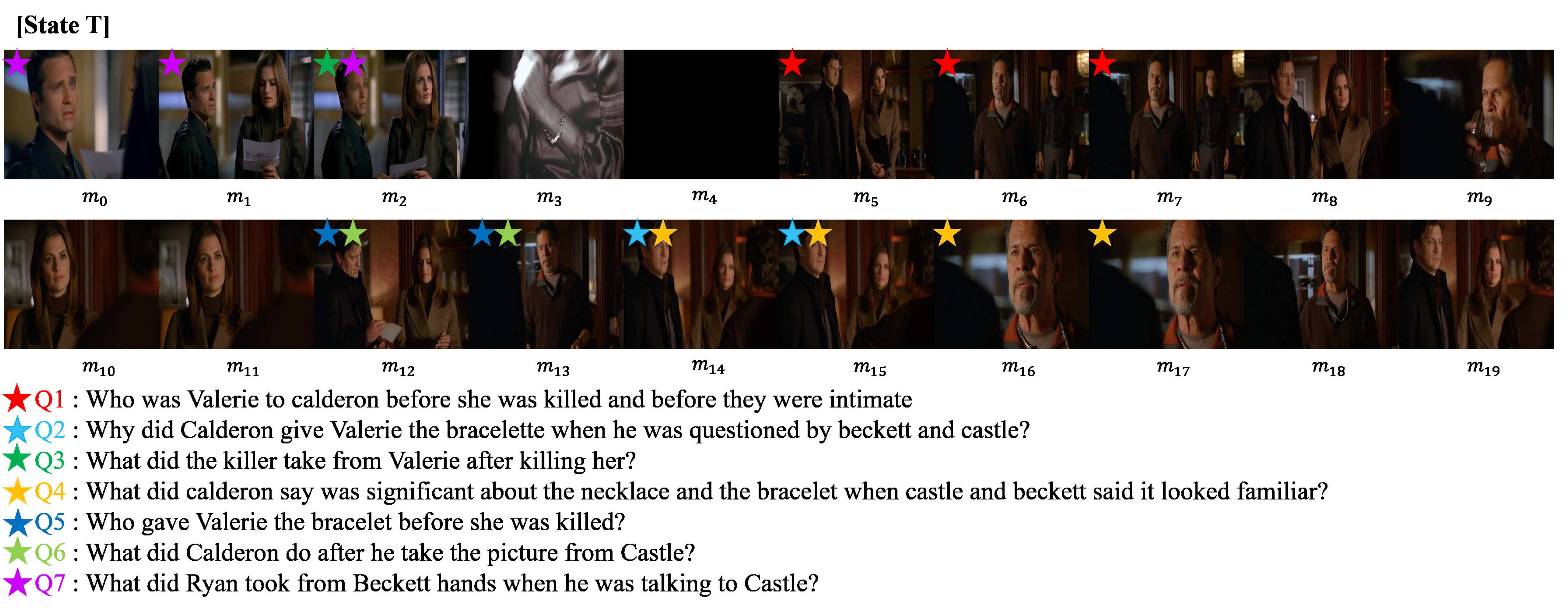} \\
		\end{tabular}
		\caption{\small An example of clip from drama 'Castle'. Each frame with star is corresponding to question with the star of same color. }
		\centering
	\end{figure*}
	
	\begin{figure*}[h!]
		\centering
		\begin{tabular}{@{}c@{}}
			\includegraphics[width=\linewidth]{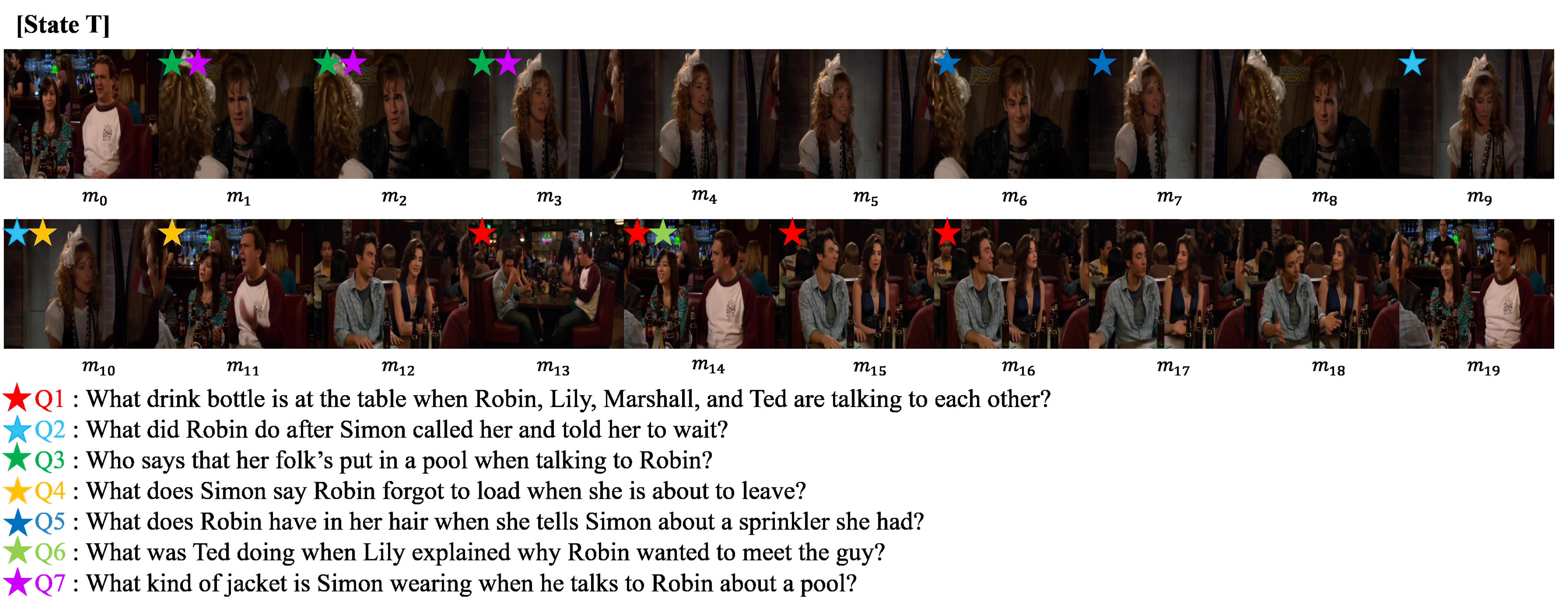} \\
		\end{tabular}
		\caption{\small An example of clip from drama 'When I met your mother'. Each frame with star is corresponding to question with the star of same color. }
		\centering
	\end{figure*}

\end{document}